\documentclass{article}
\usepackage{arxiv_preprint}

\usepackage[utf8]{inputenc}
\usepackage[T1]{fontenc}
\usepackage{amsmath,amssymb,amsthm,mathtools}
\usepackage{booktabs}
\usepackage{multirow}
\usepackage{array}
\usepackage{graphicx}
\usepackage{enumitem}
\usepackage{microtype}
\usepackage{url}
\usepackage{xcolor}
\definecolor{citeblue}{HTML}{1B5E8A}
\definecolor{linkred}{HTML}{8C1D40}
\definecolor{bratgreen}{HTML}{8ACE00}
\newcommand{\brat}{\textcolor{bratgreen}{\textbf{BRAT}}}
\usepackage[colorlinks=true,linkcolor=linkred,citecolor=citeblue,urlcolor=citeblue]{hyperref}
\makeatletter
\AtBeginDocument{%
  \let\brat@orig@hyper@linkstart\hyper@linkstart
  \let\brat@orig@hyper@linkend\hyper@linkend
  \def\hyper@linkstart#1#2{\hbox\bgroup\brat@orig@hyper@linkstart{#1}{#2}}%
  \def\hyper@linkend{\brat@orig@hyper@linkend\egroup}%
}
\makeatother

\newtheorem{theorem}{Theorem}
\newtheorem{proposition}[theorem]{Proposition}

\newcommand{\EE}{\mathbb{E}}
\newcommand{\VV}{\mathrm{Var}}
\newcommand{\Cov}{\mathrm{Cov}}
\newcommand{\KL}{D_{\mathrm{KL}}}
\newcommand{\TV}{D_{\mathrm{TV}}}
\newcommand{\R}{\mathbb{R}}
\newcommand{\Y}{\mathcal{Y}}
\newcommand{\X}{\mathcal{X}}
\newcommand{\sg}{\mathrm{sg}}


\allowdisplaybreaks[1]

\title{\raisebox{-0.20\height}{\includegraphics[height=1.35em]{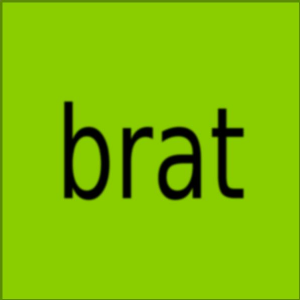}}\hspace{0.36em}BRAT: Bounded Residual-Aligned Training for Stable Multi-Pass LLM Post-Training}

\author{
\textbf{Xunlan Zhou$^{1}$ \quad
Yuzhe Zhao$^{2}$ \quad
Xuhong Huang$^{3}$ \quad
Feng Chen$^{4}$ \quad Xuanlin Chen$^{1}$} \\
\textbf{
Lang Feng$^{4}$ \quad
Bo An$^{4}$}\thanks{Corresponding author.} \\[0.5em]
$^{1}$Nanjing University \\
$^{2}$Shanghai Jiao Tong University \\
$^{3}$The University of Hong Kong \\
$^{4}$Nanyang Technological University
}

\begin{document}

\maketitle

\begin{abstract}
Reinforcement learning for large language model post-training often reuses each batch of sampled rollouts for several gradient passes to amortize the cost of generation. Under such multi-pass reuse, existing objectives fail in different ways: unclipped updates collapse as the policy drifts away from the sampling policy, adding a negative baseline avoids collapse but reinforces every sampled response, and PPO-style ratio clipping cuts off gradients abruptly. We first ask why reuse is fragile. Measuring how well an update estimated from a small rollout group aligns with the KL-regularized target on held-out rollouts of the same prompts, we find that this alignment drops by nearly an order of magnitude from in-sample to held-out rollouts across three base models, suggesting that repeated passes largely overfit the sampled group. Motivated by this finding, we propose \textbf{Bounded Residual-Aligned Training (BRAT)}, motivated by a proximal update toward a bounded reweighting of the sampling policy. BRAT assigns each token a gradient weight that decreases linearly with its probability ratio and vanishes at a trust-region boundary, and normalizes group-centered rewards by their maximum magnitude, keeping advantages zero-mean and bounded without extra hyperparameters. With four reuse passes, BRAT avoids training collapse, limits policy drift, and achieves higher average accuracy with lower seed variance on most benchmarks.
\end{abstract}
\section{Introduction}
\label{sec:intro}

Reinforcement learning with verifiable rewards (RLVR) has become a standard recipe for improving large language models (LLMs) on mathematical and algorithmic reasoning \citep{cobbe2021gsm8k,lightman2023lets,shao2024deepseekmath,liu2025understanding}. Because autoregressive generation is expensive, many LLM post-training pipelines reuse the same rollouts for several gradient updates. They sample a small group of responses per prompt from the current policy,
then make several passes over this fixed batch as the policy moves away from the one that generated it \citep{schulman2017ppo,sheng2025hybridflow}.

Reusing a finite rollout group across passes, however, exposes distinct failure modes in existing objectives (Table~\ref{tab:method_comparison}). Unclipped off-policy methods apply constant log-likelihood multipliers throughout reuse: group-centered linear updates collapse once repeated negative-advantage pushes drive off-policy token shifts unchecked, whereas Asymmetric REINFORCE \citep[AsymRE;][]{arnal2025asymre} avoids collapse by adding a negative shift ($b_{\mathrm{asym}} < 0$) to the group baseline, which adds a positive imitation term to every sampled response, including responses in all-incorrect groups, at the cost of growing actor gradients, larger variance across seeds, and lower Pass@16 at higher reuse. Conversely, ratio-clipped methods---such as GRPO \citep{shao2024deepseekmath} and Dr.GRPO \citep{liu2025understanding}---rely on PPO's hard clipping threshold \citep{schulman2017ppo} to abruptly zero gradients beyond the directional clip boundary, while recent alternatives that widen clipping bounds \citep{yu2025dapo}, taper off-policy weights \citep{leroux2025topr}, or apply ratio-space quadratic and sigmoid gates \citep[SPO, SAPO;][]{xie2025spo,gao2025sapo} still leave positive updates unattenuated, multiply gradients within the clipping interval by the probability ratio $\rho_{i,t} > 1$, or reverse gradient sign past the boundary. Independently, at the group level, GRPO divides centered rewards by the group standard deviation, which inflates rare outcomes on near-saturated prompts (almost all correct or all incorrect), whereas Dr.GRPO's unscaled centering shrinks weights to $\pm 0.5$ when half the responses are correct.

\begin{table}[t]
\centering
\caption{\textbf{Comparison of group-based post-training objectives on a rollout group of size $K$.} Here $R_i \in \{0,1\}$ denotes binary reward with group mean $\hat{p}$ and sample standard deviation $s_R$, $z_{i,t}$ is the token log-probability ratio relative to the sampling policy, $\rho_{i,t} = e^{z_{i,t}}$ is the probability ratio, and $\delta \in (0,1)$ is the trust radius.}
\label{tab:method_comparison}
\vspace{2pt}
\resizebox{\textwidth}{!}{%
\begin{tabular}{lllll}
\toprule
\textbf{Method} & \textbf{Group Advantage / Residual $A_i$} & \textbf{Advantage Range} & \textbf{Token Multiplier $m(z_{i,t}, A_i)$} & \textbf{KL reg} \\
\midrule
RWR / AWR & $e^{\eta R_i}/Z_\eta > 0$ (always positive) & $(0, e^\eta)$ & $1$ (non-saturating in $z$) & Implicit $\pi_0$ \\
GRPO & $(R_i - \hat{p}) / (s_R + 10^{-6})$ & $[-2.47, +2.47]$ ($K{=}8$) & $\rho_{i,t}\,\mathbb{I}[\text{unclipped}]$ (discontinuous) & $\beta_{\mathrm{KL}} > 0$ \\
Dr.GRPO & $R_i - \hat{p}$ & $[-0.88, +0.88]$ ($K{=}8$) & $\rho_{i,t}\,\mathbb{I}[\text{unclipped}]$ (discontinuous) & None \\
SPO & $(R_i - \hat{p}) / (s_R + 10^{-6})$ & $[-2.47, +2.47]$ ($K{=}8$) & $\rho_{i,t}\bigl(1 - \operatorname{sign}(A_i)(\rho_{i,t}-1)/\epsilon\bigr)$ (reverses) & Optional \\
SAPO & $(R_i - \hat{p}) / (s_R + 10^{-6})$ & $[-2.47, +2.47]$ ($K{=}8$) & $4\rho_{i,t}\sigma(\tau_i(\rho_{i,t}-1))\bigl(1-\sigma(\tau_i(\rho_{i,t}-1))\bigr)$ & None \\
AsymRE & $R_i - (\hat{p} + b_{\mathrm{asym}})$ ($b_{\mathrm{asym}} < 0$) & $[-1-b_{\mathrm{asym}}, 1-b_{\mathrm{asym}}]$ & $1$ (non-saturating in $z$) & None \\
\midrule
\brat\textbf{ (Ours)} & $(R_i - \hat{p}) / \max(\hat{p}, 1-\hat{p})$ & $\mathbf{[-1, +1]}$ (mixed; $0$ homog.) & $\boldsymbol{1 - \operatorname{sign}(A_i)(e^{c_{i,t}}-1)/\delta}$ (continuous, $\in [0,1]$) & \textbf{None} \\
\bottomrule
\end{tabular}%
}

\end{table}

Why do multi-pass updates on small rollout groups so easily drift? In Section~\ref{sec:geometry}, we analyze this question through the \emph{residual geometry} of KL-regularized post-training, where ideal policy improvement follows a zero-mean \emph{reward residual} that shifts probability mass from lower- to higher-reward responses. Progress per unit of policy movement toward this target is governed not by raw gradient magnitude, but by movement-normalized \emph{directional efficiency}---the cosine similarity between the reward residual and the change in log probability. By evaluating updates computed from a small rollout group ($K=16$ in-sample rollouts) on disjoint held-out rollouts of the exact same prompts across three base models, we observe a pronounced gap between in-sample and held-out alignment: directional efficiency drops by nearly ninefold from the in-sample rollouts used to estimate the update to held-out trajectories. Because directions fitted to small rollout groups overfit the sampled trajectories, repeatedly pushing the policy along a single cached group across passes rapidly accumulates off-policy drift rather than transferable policy improvement.

Guided by this diagnosis, we introduce \textbf{Bounded Residual-Aligned Training (BRAT)}. Motivated by a population forward-KL proximal projection onto a bounded linear-tilt target distribution, BRAT combines a \textbf{Directional Saturating (DS)} token objective---whose gradient multiplier decays linearly in the probability ratio from $1$ to $0$ and reaches zero at a sign-dependent trust boundary---with \textbf{Symmetric-Maximum (\texttt{SymMax})} normalization of group rewards. Within each group, the normalized weights sum to zero, minority outcomes receive unit-magnitude weights ($+1$ or $-1$), and centered rewards are scaled most strongly when the group contains equal numbers of correct and incorrect responses.

Our main contributions are threefold:
\begin{itemize}[leftmargin=1.8em,itemsep=1.5pt,topsep=2pt]
\item \textbf{Residual Geometry and In-Sample vs.\ Held-Out Alignment Diagnosis:} We characterize policy updates via movement-normalized \emph{directional efficiency} toward the signed reward residual and show across three base models that alignment drops by nearly an order of magnitude from in-sample rollouts to disjoint held-out rollouts of the same prompts, identifying in-sample overfitting of the update direction as a key bottleneck in multi-pass reuse.
\item \textbf{Bounded Residual-Aligned Training (BRAT):} Inspired by forward-KL proximal projection onto bounded linear-tilt targets, BRAT pairs a smooth \emph{Directional Saturating} token objective with parameter-free \emph{Symmetric-Maximum} (\texttt{SymMax}) group normalization, bounding every token log-ratio derivative within $[-1, 1]$ while ensuring that group weights sum to zero and assign unit-magnitude weights to minority outcomes.
\item \textbf{Multi-Pass Mechanism and Benchmark Validation:} Across multi-pass held-out alignment diagnostics, 16-round ablations, and six mathematical reasoning benchmarks, we show that BRAT suppresses off-policy drift, prevents multi-pass collapse, maintains mean competition Pass@16 under longer training, and improves accuracy while reducing variance across seeds.
\end{itemize}

\section{Background and Related Work}
\label{sec:background}

% \paragraph{KL-Regularized Post-Training and the Signed Reward Residual.}
% Given a prompt distribution $\mathcal{D}$, a reference policy $\pi_0$, a bounded reward $R(x,y) \in [0,1]$, and an inverse temperature $\eta > 0$, KL-regularized policy optimization maximizes expected reward subject to a reverse-KL penalty toward the reference policy \citep{peters2010reps,abdolmaleki2018mpo,korbak2022rl,rafailov2023dpo}. Its unique closed-form optimizer on the reference support is the exponentially tilted Gibbs distribution, which can be decomposed into the reference anchor plus a zero-sum signed residual:
% \begin{align}
% G_\eta(y\mid x) &= \frac{\pi_0(y\mid x)e^{\eta R(x,y)}}{Z_\eta(x)},
% \qquad
% Z_\eta(x) \coloneqq \EE_{y\sim\pi_0(\cdot\mid x)}\!\left[e^{\eta R(x,y)}\right],
% \label{eq:gibbs_target} \\
% \Delta_\eta(y\mid x) &\coloneqq G_\eta(y\mid x) - \pi_0(y\mid x) = \pi_0(y\mid x)a_\eta(x,y),
% \qquad
% a_\eta(x,y) \coloneqq \frac{e^{\eta R(x,y)}}{Z_\eta(x)} - 1.
% \label{eq:signed_residual}
% \end{align}
% Because the relative residual weight $a_\eta(x,y)$ has zero expectation under $\pi_0(\cdot\mid x)$, the \emph{signed reward residual} $\Delta_\eta$ sums to zero on every prompt and isolates the exact probability mass transfer required to move from $\pi_0$ to $G_\eta$.

\paragraph{KL-Regularized Post-Training and the Signed Reward Residual.}
Given a prompt distribution $\mathcal{D}$, a reference policy $\pi_0$, a bounded reward $R(x,y) \in [0,1]$, and a reward-tilt strength $\eta > 0$ (the inverse KL penalty weight), KL-regularized policy optimization maximizes $\EE_{y\sim q(\cdot\mid x)}[R(x,y)] - \frac{1}{\eta}\KL(q(\cdot\mid x)\Vert \pi_0(\cdot\mid x))$ \citep{peters2010reps,abdolmaleki2018mpo,korbak2022rl,rafailov2023dpo}. Its closed-form optimizer on the reference support is the exponentially tilted Gibbs distribution, which decomposes into the reference anchor plus a zero-sum signed residual:
\begin{align}
G_\eta(y\mid x) &= \frac{\pi_0(y\mid x)e^{\eta R(x,y)}}{Z_\eta(x)},
\qquad
Z_\eta(x) \coloneqq \EE_{y\sim\pi_0(\cdot\mid x)}\!\left[e^{\eta R(x,y)}\right],
\label{eq:gibbs_target} \\
\Delta_\eta(y\mid x) &\coloneqq G_\eta(y\mid x) - \pi_0(y\mid x) = \pi_0(y\mid x)a_\eta(x,y),
\qquad
a_\eta(x,y) \coloneqq \frac{e^{\eta R(x,y)}}{Z_\eta(x)} - 1.
\label{eq:signed_residual}
\end{align}
Because the relative residual weight $a_\eta(x,y)$ has zero expectation under $\pi_0(\cdot\mid x)$, the \emph{signed reward residual} $\Delta_\eta$ sums to zero on every prompt and isolates the exact probability mass transfer required to move from $\pi_0$ to $G_\eta$.

\paragraph{Positive-Weight Tilting vs.\ Centered Group Baselines.}
Table~\ref{tab:method_comparison} summarizes how existing post-training objectives approximate this update on a sampled group. Classical weighted-regression methods---such as RWR \citep{peters2007rwr,strupl2022rwr} and AWR \citep{peng2019awr} (with canonical exponential weights $e^{\eta R_i}/Z_\eta > 0$), as well as threshold-filtered variants such as V-MPO \citep{song2020vmpo} and ReST \citep{gulcehre2023rest,singh2024beyond}---assign non-negative weights that mix reference imitation with the directional reward signal. Critic-free RL methods instead center rewards within each prompt group, as in RLOO \citep{kool2019buy,ahmadian2024back}, GRPO \citep{shao2024deepseekmath}, and Dr.GRPO \citep{liu2025understanding}. While Dr.GRPO removes GRPO's length and standard-deviation biases by using constant length normalization and unscaled centered residuals $R_i - \hat{p}$, both methods rely on PPO's hard ratio clipping, which multiplies gradients within the clipping interval by $\rho_{i,t} = e^{z_{i,t}}$ before abruptly zeroing gradients beyond the directional clip boundary $1 \pm \epsilon$.

\paragraph{Modified Clipping, Soft Weighting Schemes, and Unclipped Alternatives.}
Classical trust-region and mirror-descent policy optimization \citep{kakade2001natural,schulman2015trpo,tomar2020mdpo,kimi2025k15} and off-policy trace-cutting methods \citep{munos2016retrace,espeholt2018impala} constrain policy updates via divergence penalties or truncated importance ratios. In LLM post-training, recent variants relax hard ratio clipping during multi-pass reuse by widening the clipping window \citep[DAPO;][]{yu2025dapo}, clipping detached importance weights \citep[CISPO;][]{chen2025cispo}, operating on sequence-level ratios \citep[GSPO;][]{zheng2025gspo}, filtering noisy token ratios \citep[KPO;][]{kpo2026}, or asymmetrically tapering off-policy weights \citep[TOPR;][]{leroux2025topr}. Smooth or structural alternatives replace hard clipping with quadratic ratio penalties \citep[SPO;][]{xie2025spo}, sigmoid soft gates \citep[SAPO;][]{gao2025sapo}, generalized Bregman mirror maps \citep[GBMPO;][]{gbmpo2026}, dual TV trust-region rollback \citep[FiberPO;][]{fiberpo2026}, listwise simplex projection \citep[LPO;][]{lpo2026}, moving reference anchors \citep{anchored2026}, or adaptive rollout reuse \citep[AR3PO;][]{ar3po2025}. Conversely, AsymRE \citep{arnal2025asymre} drops ratio clipping altogether and shifts the group baseline downward by $b_{\mathrm{asym}} < 0$; algebraically, $A_i^{\mathrm{AsymRE}} = (R_i - \hat{p}) + |b_{\mathrm{asym}}|$ adds a positive imitation term $|b_{\mathrm{asym}}|$ to every sampled response, including responses in all-incorrect groups, preventing centered linear collapse at the cost of reinforcing existing failure modes. BRAT differs from these approaches by ensuring that group weights sum to zero while introducing an advantage-signed truncated $k_3$ penalty \citep{schulman2020kl} whose token gradient multiplier decays linearly with $\rho_{i,t}$ from $1$ to $0$ at the trust boundary, avoiding both the $\rho_{i,t} > 1$ importance-ratio factor within the clipping interval of PPO and SAPO and the post-boundary gradient sign reversal of SPO.

\section{Residual Geometry and the In-Sample vs.\ Held-Out Alignment Gap}
\label{sec:geometry}

\subsection{Residual Decomposition of the Target Cross-Entropy Gradient}
\label{sec:student_geometry}

At the non-parametric distribution level, increasing the inverse temperature $\eta$ traces a standard exponential-tilting path whose reward gain and KL divergence satisfy exact variance integral identities (Proposition~\ref{prop:dist_flow} in Appendix~\ref{app:proof_dist}, verified empirically in Appendix~\ref{app:exp_ab_details}). Consider a parametric policy $\pi_\theta(y\mid x)$ with score function $s_\theta(x,y) \coloneqq \nabla_\theta \log \pi_\theta(y\mid x)$. Substituting the residual decomposition~\eqref{eq:signed_residual} into the target cross-entropy $\mathcal{L}_{G_\eta}(\theta) \coloneqq -\EE_{\mathcal{D}}\EE_{G_\eta}[\log \pi_\theta(y\mid x)]$ separates the reference cross-entropy $\mathcal{L}_0(\theta) \coloneqq -\EE_{\mathcal{D}}\EE_{\pi_0}[\log \pi_\theta(y\mid x)]$ from the centered residual objective:

\begin{proposition}[Residual Decomposition of the Target Gradient and Directional Efficiency]
\label{prop:pullback}
Define the \emph{Residual-Aligned Training (RAT)} functional and negative residual gradient:
\begin{align}
\mathcal{L}_{\mathrm{RAT}}(\theta; \eta)
&\coloneqq -\EE_{x\sim\mathcal{D}}\!\sum_{y\in\Y} \Delta_\eta(y\mid x)\log \pi_\theta(y\mid x)
= -\EE_{x\sim\mathcal{D}}\EE_{y\sim\pi_0}\!\left[a_\eta(x,y)\log \pi_\theta(y\mid x)\right],
\label{eq:rat_loss} \\
g_\eta(\theta)
&\coloneqq -\nabla_\theta \mathcal{L}_{\mathrm{RAT}}(\theta; \eta)
= \EE_{x\sim\mathcal{D}}\EE_{y\sim\pi_0}\!\left[a_\eta(x,y)\,s_\theta(x,y)\right].
\label{eq:rat_grad}
\end{align}
Then $\mathcal{L}_{G_\eta}(\theta) = \mathcal{L}_0(\theta) + \mathcal{L}_{\mathrm{RAT}}(\theta; \eta)$, and at $\pi_\theta = \pi_0$ the reference gradient vanishes so $g_\eta(\theta) = -\nabla_\theta \mathcal{L}_{G_\eta}(\theta)$. For any update direction $d$, let $\delta_\theta(x,y; d) \coloneqq s_\theta(x,y)^\top d$ denote the linearized log-probability shift. Under the joint expectation $\EE_{\pi_0}[\cdot]$ over $x\sim\mathcal{D}$ and $Y\sim\pi_0(\cdot\mid x)$, first-order residual improvement satisfies
\begin{align}
\langle g_\eta(\theta), d\rangle
&= \sqrt{\EE_{\pi_0}\!\left[a_\eta^2\right]\EE_{\pi_0}\!\left[\delta_\theta^2\right]}\;\rho_\eta^{\mathrm{pred}}(\theta; d),
\label{eq:directional_efficiency} \\
\rho_\eta^{\mathrm{pred}}(\theta; d)
&\coloneqq \frac{\EE_{\pi_0}\!\left[a_\eta(x,Y)\,\delta_\theta(x,Y; d)\right]}{\sqrt{\EE_{\pi_0}\!\left[a_\eta(x,Y)^2\right]\EE_{\pi_0}\!\left[\delta_\theta(x,Y; d)^2\right]}} \in [-1, 1].
\label{eq:rho_pred_def}
\end{align}
\end{proposition}

Proposition~\ref{prop:pullback} (proved in Appendix~\ref{app:proof_student}) separates the first-order predicted gain $\langle g_\eta(\theta), d\rangle$ from movement-normalized \emph{directional efficiency} $\rho_\eta^{\mathrm{pred}}$---the cosine similarity between the reward residual and the change in log probability (with finite-step realized efficiency $\rho_\eta^{\mathrm{obs}}$ defined via odd central-difference log-probability shifts).

\subsection{Empirical Diagnosis: In-Sample vs.\ Held-Out Alignment}
\label{sec:exp_b}

\begin{figure}[!t]
\centering
\includegraphics[width=0.92\textwidth]{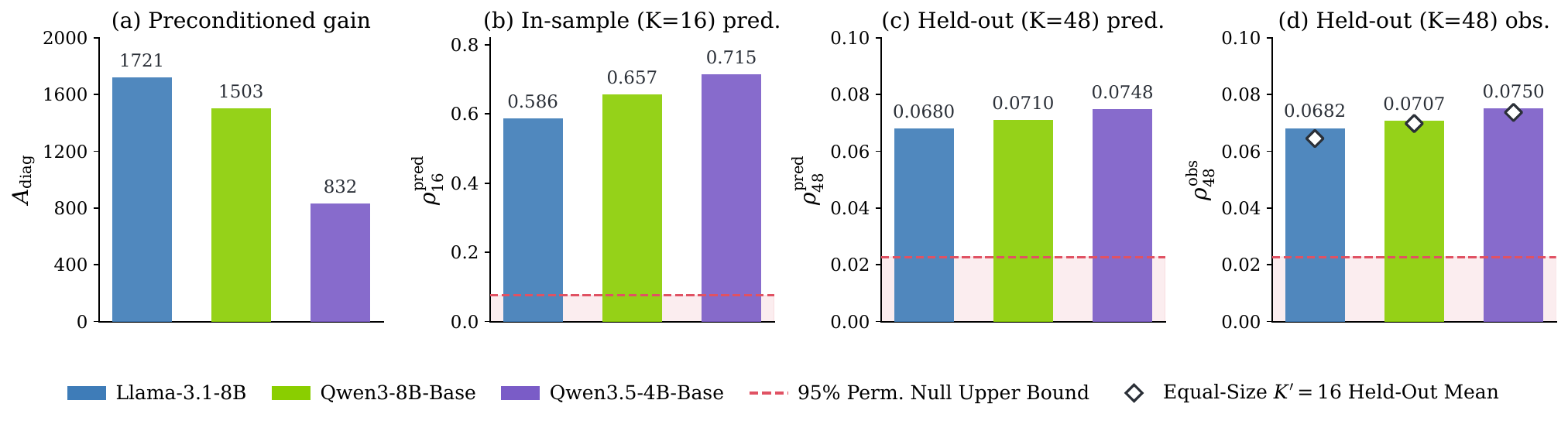}
\vspace{-5pt}
\caption{\textbf{In-sample vs.\ held-out residual alignment diagnostic across three base models ($128$ prompts, $\eta=1.0$).} \textbf{(a)} First-order preconditioned gain $A_{\mathrm{diag}} = \langle \hat{g}_{\eta,16}, \hat{d}\rangle$ on the $K=16$ in-sample split. \textbf{(b)} In-sample directional efficiency $\rho_{16}^{\mathrm{pred}}$ vs.\ the $K=16$ permutation null ($95\%$ upper bound, dashed red). \textbf{(c)} Held-out linearized directional efficiency $\rho_{48}^{\mathrm{pred}}$ on $48$ disjoint rollouts of the same prompts. \textbf{(d)} Held-out realized directional efficiency $\rho_{48}^{\mathrm{obs}}$ on $K=48$ (bars), mean efficiency across three equal-size $K'=16$ held-out blocks (diamonds), and the $K=48$ $95\%$ permutation null upper bound (dashed red).}
\label{fig:exp_b_main}
\vspace{-4pt}
\end{figure}

\paragraph{The Gap Between In-Sample and Held-Out Alignment.}
To test how well a small-group update transfers to new responses of the same prompt, we evaluate Proposition~\ref{prop:pullback} at $\eta = 1.0$ across three base models---\texttt{Llama-3.1-8B} \citep{llama3.1}, \texttt{Qwen3-8B-Base} \citep{yang2025qwen3}, and \texttt{Qwen3.5-4B-Base} \citep{qwen35}---on $128$ MATH prompts using a shared set of $64$ cached reference rollouts per prompt, split into $K=16$ rollouts used to estimate the update (in-sample rollouts) and $K=48$ disjoint held-out rollouts. On the $K=16$ in-sample split, we compute the empirical residual gradient $\hat{g}_{\eta,16}$ in~\eqref{eq:rat_grad} and take a diagonally preconditioned update step $d = \hat{d}$, comparing the unnormalized first-order gain $A_{\mathrm{diag}} \coloneqq \langle \hat{g}_{\eta,16}, \hat{d}\rangle$---the left-hand side of~\eqref{eq:directional_efficiency}---with directional efficiency on both the $K=16$ in-sample split ($\rho_{16}^{\mathrm{pred}}$) and the $K=48$ held-out split ($\rho_{48}^{\mathrm{pred}}, \rho_{48}^{\mathrm{obs}}$; full protocol in Appendix~\ref{app:exp_ab_details}, Table~\ref{tab:app_exp_b}). Figure~\ref{fig:exp_b_main} shows two patterns. First, unnormalized gain $A_{\mathrm{diag}}$ varies inversely with held-out directional efficiency across the three architectures (Figure~\ref{fig:exp_b_main}a vs.\ d), showing that raw inner-product gain is confounded by parameter dimension and step scaling. Second, across all three models, \textbf{directional efficiency drops by nearly $9\times$ when moving from the $K=16$ in-sample rollouts to the $K=48$ held-out rollouts of the exact same prompts}, retaining only about $11\%$ of its in-sample alignment (Figure~\ref{fig:exp_b_main}b vs.\ c--d). Two controls in Figure~\ref{fig:exp_b_main}d and Table~\ref{tab:app_exp_b} rule out sample-size and random-noise artifacts: partitioning the held-out rollouts into disjoint equal-size blocks of $K'=16$ responses preserves the same ninefold drop across all three models, while within-prompt reward permutation tests confirm that held-out directional efficiency remains $2.6$ to $3.9$ times above the $95\%$ permutation null threshold ($5.4\text{--}7.7$ standard deviations above chance). Together with the close agreement between linearized and finite-step efficiency (Figure~\ref{fig:exp_b_main}c--d), this shows that an update direction fitted to $K=16$ rollouts in a high-dimensional parameter space ($\dim(\theta) \gg K$) combines a statistically significant cross-trajectory reward signal with an eightfold larger in-sample overfitting component.

\paragraph{From In-Sample Overfitting to Bounded Multi-Pass Updates.}
Consider a multi-pass update from the sampling policy $q$ to a candidate policy $p$, which changes sequence log-probabilities by $u(x,y) \coloneqq \log(p(y\mid x)/q(y\mid x))$. To separate the magnitude of this update from its direction, let $M$ denote the RMS log-probability drift, $\hat{u}$ the normalized shift direction ($\EE_q[\hat{u}^2]=1$), $a_{\eta,q}(x,y) \coloneqq e^{\eta R(x,y)}/\EE_{Y\sim q(\cdot\mid x)}[e^{\eta R(x,Y)}] - 1$ the relative residual under $q$, and $\Gamma(\hat{u})$ the gap between in-sample alignment on the sampled group and population alignment under $q$:
\begin{align}
M \coloneqq \sqrt{\EE_q\!\left[u(x,Y)^2\right]},
\qquad
\hat{u}(x,y) &\coloneqq \frac{u(x,y)}{M},
\qquad
\Gamma(\hat{u}) \coloneqq \widehat{\EE}_{\mathrm{train}}\!\left[a_{\eta,q}\hat{u}\right] - \EE_q\!\left[a_{\eta,q}\hat{u}\right],
\nonumber \\
\EE_q\bigl[a_{\eta,q}(x,Y)\,u(x,Y)\bigr]
&= M\Bigl(\widehat{\EE}_{\mathrm{train}}\bigl[a_{\eta,q}\hat{u}\bigr] - \Gamma(\hat{u})\Bigr).
\label{eq:generalization_bridge}
\end{align}
Equation~\eqref{eq:generalization_bridge} is an exact decomposition that clarifies why unregularized multi-pass reuse is fragile: when an update direction $\hat{u}$ estimated from a small rollout group has a large gap $\Gamma(\hat{u})$ between in-sample and held-out alignment (Figure~\ref{fig:exp_b_main}), allowing repeated off-policy passes to accumulate large sequence drift $M$ directly magnifies sample-specific overfitting relative to population improvement. This motivates our algorithmic design in Section~\ref{sec:brat}, which smoothly attenuates off-policy token log-ratio gradients as trajectories drift from $q$ while keeping group weights zero-sum and bounded.

\section{Bounded Residual-Aligned Training (BRAT)}
\label{sec:brat}

\subsection{A Proximal Objective with a Bounded Target}
\label{sec:brat_population}

Consider online round $t$ with sampling policy $q = \pi_t$. For each prompt $x$, we scale the centered reward by $\alpha(x) \ge 0$ so that the resulting residual $A_q(x,y)$ has zero mean under $q$ and is bounded in $[-1, 1]$:
\begin{equation}
A_q(x,y) \coloneqq \alpha(x)\bigl(R(x,y) - \EE_{Y\sim q(\cdot\mid x)}[R(x,Y)]\bigr) \in [-1, 1].
\end{equation}
Given a trust radius $\delta_t \in (0, 1)$, instead of targeting an unbounded exponential tilt, we define the \emph{bounded linear-tilt target distribution} $H_t$ and update the candidate policy $p$ by minimizing the forward-KL proximal objective $\mathcal{L}_t(p)$:
\begin{align}
H_t(y\mid x) &\coloneqq q(y\mid x)\bigl(1 + \delta_t A_q(x,y)\bigr),
\label{eq:linear_tilt_target} \\
\mathcal{L}_t(p) &\coloneqq \EE_{x\sim\mathcal{D}}\!\left[
-\EE_{y\sim q(\cdot\mid x)}\!\left[A_q(x,y)\log\frac{p(y\mid x)}{q(y\mid x)}\right]
+ \frac{1}{\delta_t}\KL\bigl(q(\cdot\mid x)\,\Vert\,p(\cdot\mid x)\bigr)
\right].
\label{eq:pop_brat_loss}
\end{align}
Because $\EE_q[A_q\mid x] = 0$ and $|A_q| \le 1$, $H_t$ is a valid probability distribution without a partition function, and its density ratio $H_t/q$ is strictly bounded in $[1-\delta_t, 1+\delta_t]$.

\begin{proposition}[Forward-KL Variational Projection and Multi-Round Improvement]
\label{prop:proximal_identity}
Define the scaled expected reward $J(p) \coloneqq \EE_{x\sim\mathcal{D}}\EE_{y\sim p}[\alpha(x)R(x,y)]$ and scaled reward variance $V(q) \coloneqq \EE_{x\sim\mathcal{D}}[\alpha(x)\VV_q(R\mid x)]$. For any candidate policy $p$ and trust radius $\delta_t \in (0,1)$ (proof in Appendix~\ref{app:proof_brat}):
\begin{enumerate}[leftmargin=1.8em,itemsep=1pt,topsep=1.5pt]
\item \textbf{Variational Identity:} $\delta_t \mathcal{L}_t(p) = \EE_{\mathcal{D}}\big[\KL(H_t \Vert p) - \KL(H_t \Vert q)\big]$, so $p^\star = H_t$ is the unique minimizer of $\mathcal{L}_t(p)$ on $\mathrm{supp}(q)$.
\item \textbf{Monotonic Improvement:} Whenever $\mathrm{supp}(p(\cdot\mid x)) \subseteq \mathrm{supp}(q(\cdot\mid x))$, $J(p) - J(q) \ge -\mathcal{L}_t(p)$; thus any update with $\mathcal{L}_t(p) < 0$ strictly improves $J(p)$ over $J(q)$.
\item \textbf{Multi-Round Composition:} If $e_t \coloneqq \EE_{\mathcal{D}}[\TV(\pi_{t+1}, H_t)]$, then over $T$ rounds,
\begin{equation}
\label{eq:multi_round_bound}
\EE_{x\sim\mathcal{D}}\!\left[\EE_{\pi_T}[R] - \EE_{\pi_0}[R]\right]
\ge \sum_{t=0}^{T-1}\delta_t V(\pi_t) - \sum_{t=0}^{T-1} e_t.
\end{equation}
\end{enumerate}
\end{proposition}

\subsection{Symmetric-Maximum (\texttt{SymMax}) Group Normalization}
\label{sec:symmax}

Given a group of $K$ rollouts with binary rewards $R_i \in \{0,1\}$ and empirical success rate $\hat{p} \coloneqq \frac{1}{K}\sum_{j=1}^K R_j$, Section~\ref{sec:brat_population} requires choosing a prompt-level scaling $\alpha(\hat{p}) \ge 0$ so that the empirical weights $A_i = \alpha(\hat{p})(R_i - \hat{p})$ sum to zero within each group and lie in $[-1,1]$. Existing baselines either distort rare outcomes or downweight balanced groups (Table~\ref{tab:method_comparison}): GRPO uses $\alpha(\hat{p}) = 1/s_R \propto 1/\sqrt{\hat{p}(1-\hat{p})}$, which amplifies rare outcomes on near-saturated prompts (almost all correct or all incorrect); Dr.GRPO uses unscaled centering $\alpha(\hat{p}) = 1$, which shrinks weights to $\pm 0.5$ when half the responses are correct ($\hat{p}=0.5$); and AsymRE subtracts a negative shift $b_{\mathrm{asym}} < 0$, adding a positive likelihood weight to every sampled response.

To ensure that group weights sum to zero and lie in $[-1,1]$ without hyperparameters, BRAT divides centered rewards by their maximum magnitude $\max(\hat{p}, 1-\hat{p})$ on each mixed group, defining the \textbf{Symmetric-Maximum (\texttt{SymMax})} weight with prompt-level scaling factor $\alpha(\hat{p}) = 1/\max(\hat{p}, 1-\hat{p}) \in [\frac{K}{K-1}, 2]$:
\begin{equation}
\label{eq:symmax_def}
A_i^{\mathrm{BRAT}} \coloneqq \frac{R_i - \hat{p}}{\max(\hat{p}, 1-\hat{p})}\,\mathbb{I}_{0 < \hat{p} < 1},
\quad
A_+^{\mathrm{BRAT}} = \min\!\left(1, \tfrac{1-\hat{p}}{\hat{p}}\right),
\quad
A_-^{\mathrm{BRAT}} = -\min\!\left(1, \tfrac{\hat{p}}{1-\hat{p}}\right).
\end{equation}
On mixed groups ($0 < \hat{p} < 1$), the weights sum to zero and minority outcomes receive a weight of $+1$ (when $0 < \hat{p} \le 0.5$) or $-1$ (when $0.5 \le \hat{p} < 1$). Relative to unscaled centering ($R_i - \hat{p}$), the scaling factor $\alpha(\hat{p})$ is largest ($\alpha=2$) when the group contains equal numbers of correct and incorrect responses ($\hat{p}=0.5$) and approaches $1$ ($\frac{K}{K-1}$) on near-saturated groups; thus it is not equivalent to a global learning-rate multiplier, which would over-step on near-saturated prompts.

\subsection{Token-Level Directional Saturating (DS) Surrogate}
\label{sec:ds_barrier}

Proposition~\ref{prop:proximal_identity} characterizes an idealized sequence-level projection onto the linearized tilt target $H_t(y\mid x) = q(y\mid x)(1 + \delta_t A_q(x,y))$, whereas autoregressive LLMs are updated token by token across shared prefixes. Let $z_{i,t}(\theta) \coloneqq \log\frac{\pi_\theta(y_{i,t}\mid x, y_{i,<t})}{q(y_{i,t}\mid x, y_{i,<t})}$ denote the token log-probability ratio relative to the sampling policy $q$. At each prefix, Schulman's $k_3$ KL estimator $e^{z_{i,t}} - 1 - z_{i,t} \ge 0$ \citep{schulman2020kl} is a non-negative, unbiased single-token estimator of $\KL(q \Vert \pi_\theta)$. However, applying an unweighted, unclamped $k_3$ penalty to every token would penalize zero-advantage homogeneous groups ($A_i^{\mathrm{BRAT}}=0$) and create steep exponential penalties whenever shared parameters move a token log-ratio opposite to the advantage sign ($\operatorname{sign}(A_i^{\mathrm{BRAT}})z_{i,t} < 0$).

\begin{figure}[!t]
\centering
\includegraphics[width=0.92\textwidth]{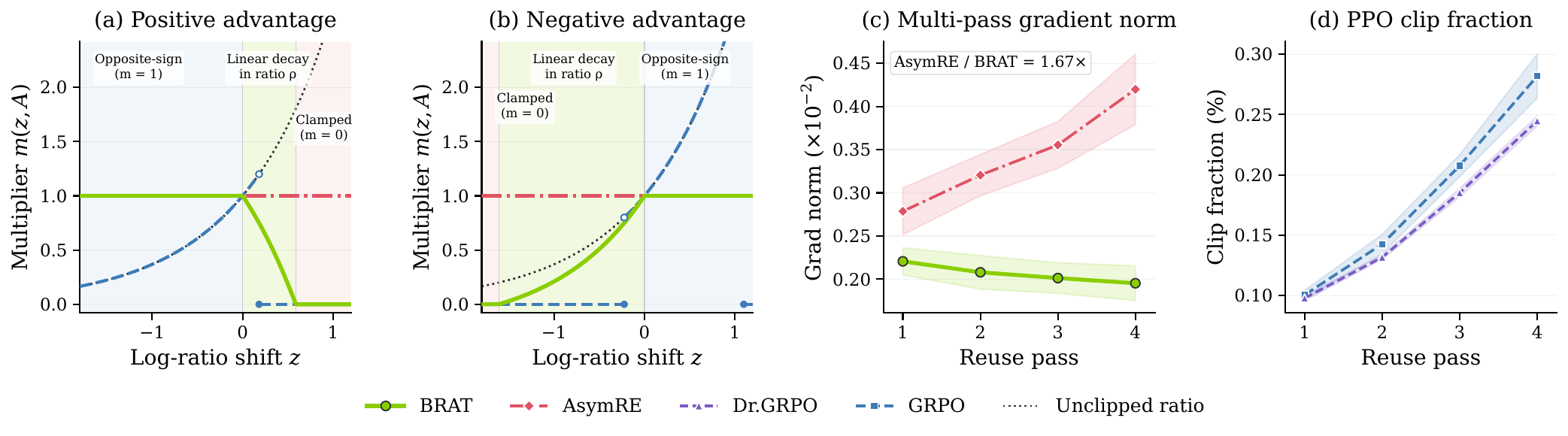}
\vspace{-5pt}
\caption{\textbf{BRAT token gradient multiplier and pass-wise training diagnostics.} \textbf{(a, b)} Token gradient multiplier vs.\ token log-probability ratio $z = \log\rho$ for positive ($A>0$) and negative ($A<0$) residuals: BRAT decays linearly with $\rho = e^z$ from $1$ to $0$ across the trust region, whereas AsymRE stays constant at $1$ and GRPO/Dr.GRPO drop discontinuously at the clip boundary. \textbf{(c)} Actor gradient norm across off-policy passes $s=1\dots 4$ ($\text{mean} \pm \text{SD}$ over 3 seeds): BRAT's gradient norm decreased from Pass~1 to Pass~4 while AsymRE's increased. \textbf{(d)} Fraction of hard-clipped tokens for GRPO and Dr.GRPO across passes.}
\label{fig:brat_geometry}
\vspace{-4pt}
\end{figure}

To construct a tractable token-level surrogate inspired by~\eqref{eq:pop_brat_loss}, we weight the truncated $k_3$ penalty by $|A_i^{\mathrm{BRAT}}|$ and restrict its exponential curvature to the \emph{directional trust interval} $I_\delta(A_i^{\mathrm{BRAT}})$ between zero shift and the directional ratio boundary $\rho_{i,t} = e^{z_{i,t}} = 1 + \delta\operatorname{sign}(A_i^{\mathrm{BRAT}})$:
\begin{align}
c_{i,t}(\theta) &\coloneqq \operatorname{clamp}\!\bigl(z_{i,t}(\theta),\, I_\delta(A_i^{\mathrm{BRAT}})\bigr),
\label{eq:directional_bounds} \\
I_\delta(A) &\coloneqq \bigl[0,\, \ln(1+\delta)\bigr]\;\text{if } A>0,
\qquad
\bigl[\ln(1-\delta),\, 0\bigr]\;\text{if } A<0,
\qquad
[0,\,0]\;\text{if } A=0.
\label{eq:directional_interval}
\end{align}
Outside $I_\delta(A_i^{\mathrm{BRAT}})$, we replace the exponential $k_3$ curve with its first-order tangent using the stop-gradient operator $\sg(\cdot)$, keeping the surrogate $C^1$-smooth across the boundary. Summing over tokens with constant length normalization $C$ \citep{liu2025understanding} over a mini-batch of sampled trajectories $\mathcal{B}$ yields the \textbf{BRAT token loss} and mini-batch objective:
\begin{align}
\psi_{\mathrm{DS}}(z, c) &\coloneqq (e^c - 1 - c) + (z - c)(e^{\sg(c)} - 1),
\label{eq:psi_ds} \\
\ell_{i,t}^{\mathrm{BRAT}}(\theta)
&\coloneqq -A_i^{\mathrm{BRAT}} z_{i,t}(\theta) + \frac{|A_i^{\mathrm{BRAT}}|}{\delta}\psi_{\mathrm{DS}}(z_{i,t}, c_{i,t}),
\label{eq:brat_token_loss} \\
\widehat{\mathcal{L}}_{\mathrm{BRAT}}(\theta)
&\coloneqq \frac{1}{|\mathcal{B}|}\sum_{i\in\mathcal{B}} \frac{1}{C}\sum_{t=1}^{|y_i|} \ell_{i,t}^{\mathrm{BRAT}}(\theta).
\label{eq:brat_minibatch_loss}
\end{align}

\begin{proposition}[Continuous and Bounded Directional Gradient Multiplier]
\label{prop:ds_gradient}
For any $\delta \in (0,1)$, $\ell_{i,t}^{\mathrm{BRAT}}(\theta)$ is globally $C^1$-smooth in $z_{i,t}$, and its log-ratio derivative satisfies (proof in Appendix~\ref{app:proof_brat}):
\begin{align}
\frac{\partial \ell_{i,t}^{\mathrm{BRAT}}}{\partial z_{i,t}}
&= -A_i^{\mathrm{BRAT}}\,m_{\mathrm{BRAT}}\bigl(z_{i,t}, A_i^{\mathrm{BRAT}}\bigr),
\qquad
\left|\frac{\partial \ell_{i,t}^{\mathrm{BRAT}}}{\partial z_{i,t}}\right| \le \bigl|A_i^{\mathrm{BRAT}}\bigr| \le 1,
\label{eq:brat_multiplier} \\
m_{\mathrm{BRAT}}(z, A) &\coloneqq 1 - \frac{\operatorname{sign}(A)\bigl(e^{c(z,A)} - 1\bigr)}{\delta}
= \operatorname{clip}\!\left(1 - \frac{\operatorname{sign}(A)(e^z - 1)}{\delta},\, 0,\, 1\right) \in [0, 1].
\label{eq:brat_multiplier_def}
\end{align}
\end{proposition}
As shown in Figure~\ref{fig:brat_geometry}a--b, the multiplier $m_{\mathrm{BRAT}}$ is continuous across three regions: (i) when a token log-ratio moves opposite to the advantage sign ($\operatorname{sign}(A)z \le 0$), the multiplier stays at $1$; (ii) inside the trust interval $I_\delta(A)$, it decays linearly with the probability ratio $\rho=e^z$ from $1$ down to $0$; and (iii) beyond the trust boundary $\rho = 1\pm\delta$, it remains at $0$ so that already-shifted tokens receive no further gradient push.

\paragraph{Relation to the Population Target and Comparing Integrated Gradient Multipliers.}
Factoring $|A_i^{\mathrm{BRAT}}|$ outside both terms in~\eqref{eq:brat_token_loss} separates group-level scaling from token-level attenuation: for minority outcomes ($|A_i^{\mathrm{BRAT}}|=1$), the boundary $1 + \delta\operatorname{sign}(A_i^{\mathrm{BRAT}}) = 1\pm\delta$ matches the sequence linear-tilt ratio $1 + \delta A_i^{\mathrm{BRAT}}$, whereas majority outcomes ($0 < |A_i^{\mathrm{BRAT}}| < 1$) share the same trust boundary $1\pm\delta$ with step magnitudes scaled proportionally by $|A_i^{\mathrm{BRAT}}|$ (keeping group updates zero-sum at $z_{i,t}=0$). Unlike PPO, SPO, or SAPO---which retain the off-policy importance ratio $\rho = e^z$ and thus weight gradients within the clipping interval by $\rho > 1$ when $z > 0$ (or reverse gradient sign past $1+\epsilon$ in SPO)---DS omits the $\rho$ factor in favor of a biased, attenuated update whose gradient multiplier $1 - \operatorname{sign}(A)(e^z-1)/\delta$ decreases with $\rho$ so that under gradient flow, token log-ratios approach the trust boundary smoothly. Because $m_{\mathrm{BRAT}}$ decays from $\rho=1$, comparing the integrated gradient multipliers over $z \in [0, \ln(1+\delta)]$ vs.\ PPO's $I_{\mathrm{PPO}}^+(\epsilon) = \int_0^{\ln(1+\epsilon)} e^z\,dz = \epsilon$ gives
\begin{equation}
\label{eq:ds_impulse}
I_{\mathrm{DS}}^+(\delta) \coloneqq \int_0^{\ln(1+\delta)}\!\left(1 - \frac{e^z - 1}{\delta}\right)dz = \left(1 + \frac{1}{\delta}\right)\ln(1+\delta) - 1.
\end{equation}
Thus $\delta = 0.2$ yields less than half of PPO's integrated multiplier at $\epsilon=0.2$ ($I_{\mathrm{DS}}^+(0.2) = 0.0939$ vs.\ $0.200$), whereas $\delta = 0.8$ ($I_{\mathrm{DS}}^+(0.8) = 0.3225$) matches an effective clipping radius $\epsilon_{\mathrm{eff}} \approx 0.32$. While Proposition~\ref{prop:ds_gradient} bounds scalar token log-ratio derivatives rather than full parameter-gradient norms or sequence KL (Appendix~\ref{app:proof_brat}), implementing BRAT requires only elementary tensor-wise clamping with no measurable wall-clock overhead relative to standard clipping (Appendix~\ref{app:implementation} and Table~\ref{tab:app_length_time}).

\section{Experiments}
\label{sec:experiments}

\subsection{Experimental Setup}
\label{sec:exp_setup}

\paragraph{Training Protocol and Settings.}
We post-train \texttt{Qwen3-8B-Base} \citep{yang2025qwen3} with LoRA ($r=32, \alpha=32$) \citep{hu2022lora} on verified problems from \texttt{NuminaMath-1.5} \citep{li2024numinamath}. Each online round samples $M=64$ prompts with $K=8$ rollouts, recomputes reference log-probabilities, and performs $p=4$ off-policy reuse passes ($32$ mini-batch steps per round) under constant length normalization \citep{liu2025understanding} across $3$ random seeds (Table~\ref{tab:app_hyperparams} in Appendix~\ref{app:implementation}). Experiments comprise two hardware-matched settings: a \textbf{single-GPU 12-round main setting} ($384$ updates) comparing \textbf{BRAT} against \textbf{GRPO} \citep{shao2024deepseekmath}, \textbf{Dr.GRPO} \citep{liu2025understanding}, and \textbf{AsymRE} \citep{arnal2025asymre} on accuracy, variance across seeds, and wall-clock speed (Table~\ref{tab:main_results}; Tables~\ref{tab:app_per_seed}--\ref{tab:app_length_time}), and a \textbf{distributed 8-GPU 16-round ablation suite} ($512$ updates) evaluating component factorials, reuse depth, trust radii, and token-level weighting schemes (Tables~\ref{tab:app_online_ablations}--\ref{tab:app_token_gates}).

\paragraph{Benchmarks and Metrics.}
We evaluate Greedy Pass@1 ($\text{temperature}=0$) on six mathematical reasoning benchmarks ($1{,}546$ problems): \textbf{MATH-500} \citep{lightman2023lets}, \textbf{OlympiadBench} \citep{he2024olympiadbench}, \textbf{Minerva-Math} \citep{lewkowycz2022minerva}, \textbf{AMC 2023}, \textbf{AIME 2024}, and \textbf{AIME 2025}, plus stochastic multi-sample accuracy ($\text{temperature}=1.0$) on the three competition benchmarks via Pass@4 and 16-sample competition Avg@16 and Pass@16.

\subsection{Main Benchmark Results and Cross-Seed Stability}
\label{sec:main_results}

\begin{table}[!t]
\centering
\caption{\textbf{Main benchmark evaluation (Greedy Pass@1, \%) across six mathematical reasoning tasks on \texttt{Qwen3-8B-Base}.} All methods are trained in the single-GPU 12-round, 4-pass main setting ($384$ updates) and reported as $\text{mean} \pm \text{SD}$ across 3 random seeds. \textbf{Bold} marks the highest mean score on each row.}
\label{tab:main_results}
\vspace{2pt}
\small
\setlength{\tabcolsep}{3.5pt}
\begin{tabular}{lcccccc}
\toprule
\textbf{Benchmark} & \textbf{Problems} & \textbf{Base (Init)} & \textbf{GRPO} & \textbf{Dr.GRPO} & \textbf{AsymRE} & \brat\textbf{ (Ours)} \\
\midrule
MATH-500 & 500 & 61.20 & $77.93 \pm 1.47$ & $78.20 \pm 0.69$ & $78.60 \pm 0.69$ & $\mathbf{79.60 \pm 0.40}$ \\
OlympiadBench & 674 & 26.56 & $41.39 \pm 2.14$ & $41.20 \pm 2.27$ & $41.64 \pm 4.55$ & $\mathbf{42.83 \pm 0.56}$ \\
Minerva-Math & 272 & 21.32 & $28.55 \pm 3.13$ & $25.74 \pm 0.37$ & $\mathbf{30.02 \pm 2.81}$ & $28.92 \pm 1.06$ \\
AMC 2023 & 40 & 30.00 & $53.33 \pm 2.89$ & $54.17 \pm 3.82$ & $60.00 \pm 7.50$ & $\mathbf{61.67 \pm 1.44}$ \\
AIME 2024 & 30 & 0.00 & $8.89 \pm 5.09$ & $13.33 \pm 3.33$ & $14.44 \pm 3.85$ & $\mathbf{15.56 \pm 1.92}$ \\
AIME 2025 & 30 & 6.67 & $12.22 \pm 3.85$ & $15.56 \pm 5.09$ & $8.89 \pm 1.92$ & $\mathbf{16.67 \pm 3.33}$ \\
\midrule
\textbf{Macro-Avg6} & \emph{Unweighted} & 24.29 & $37.05 \pm 1.47$ & $38.03 \pm 2.20$ & $38.93 \pm 1.07$ & $\mathbf{40.87 \pm 1.13}$ \\
\textbf{Micro-Avg} & \emph{1,546 probs} & 36.03 & $50.06 \pm 1.62$ & $49.74 \pm 1.10$ & $50.86 \pm 2.57$ & $\mathbf{51.72 \pm 0.64}$ \\
\bottomrule
\end{tabular}
\end{table}

\begin{figure}[!t]
\centering
\includegraphics[width=0.92\textwidth]{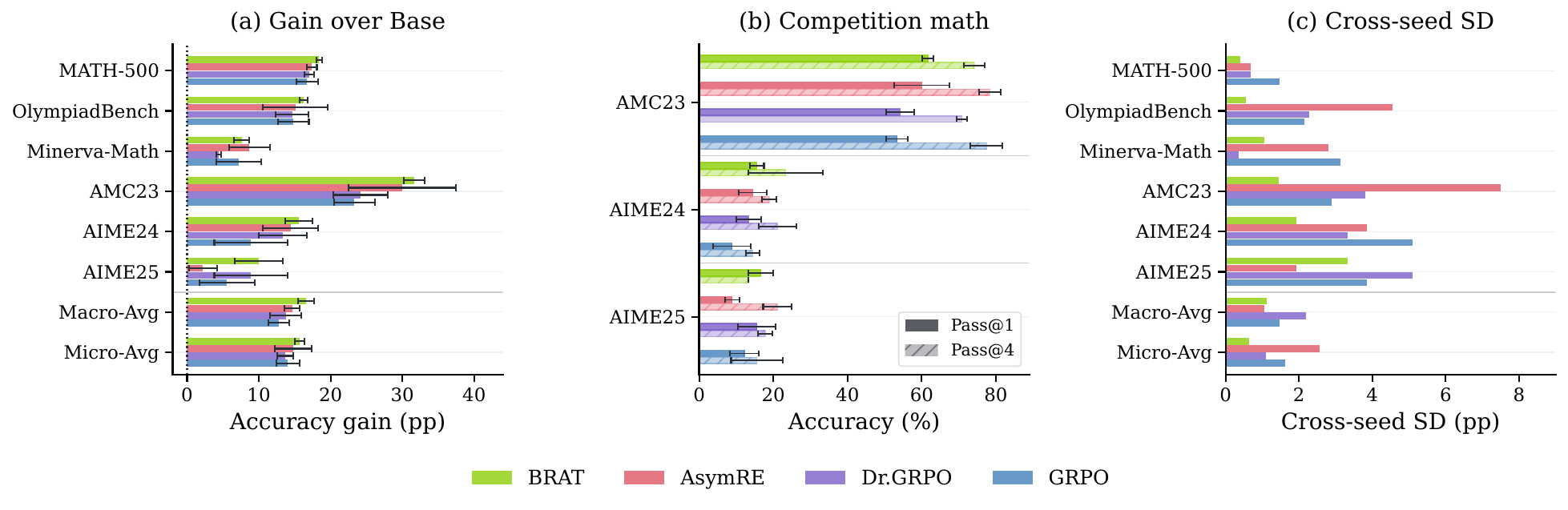}
\vspace{-5pt}
\caption{\textbf{Matched-budget benchmark performance and standard deviation across seeds on \texttt{Qwen3-8B-Base} (3 seeds, 12-round main setting).} \textbf{(a)} Greedy Pass@1 gain over Base. \textbf{(b)} Competition math Greedy Pass@1 (solid) and sampled Pass@4 (hatched). \textbf{(c)} Cross-seed standard deviation across benchmarks and averages.}
\label{fig:brat_main_perf}
\vspace{-4pt}
\end{figure}

Table~\ref{tab:main_results} and Figure~\ref{fig:brat_main_perf} summarize the 12-round main setting. Under the shared four-pass budget, BRAT achieves the highest mean greedy accuracy on five of six benchmarks and on both Macro-Avg6 ($40.87\%$) and Micro-Avg ($51.72\%$). While AsymRE attains a higher mean on Minerva-Math and higher temperature-$1.0$ Pass@4 on AMC23 and AIME25 (accompanied by shorter responses and lower truncation in Table~\ref{tab:app_length_time}), its unclipped updates exhibit larger seed-to-seed fluctuations, with SD reaching $4.55$~pp on OlympiadBench and $7.50$~pp on AMC23 (Figure~\ref{fig:brat_main_perf}c). Conversely, BRAT leads all three competition tasks on greedy Pass@1 ($31.30\%$ vs.\ AsymRE's $27.78\%$ mean) and achieves the lowest cross-seed SD on four of six tasks.

\subsection{Multi-Pass Optimization Diagnostics and Component Ablations}
\label{sec:diagnostics_stress}

\begin{figure}[!t]
\centering
\includegraphics[width=0.94\textwidth]{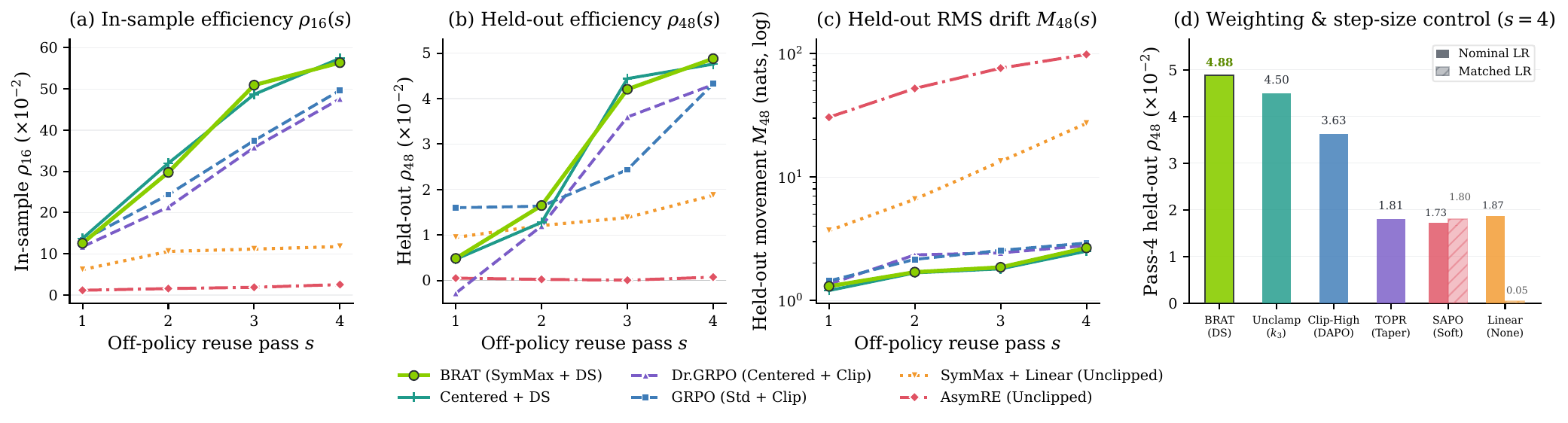}
\vspace{-5pt}
\caption{\textbf{Held-out alignment across reuse passes on \texttt{Qwen3-8B-Base} ($128$ prompts, $\eta = 0.8$):} \textbf{(a)} In-sample directional efficiency $\rho_{16}(s)$. \textbf{(b)} Held-out directional efficiency $\rho_{48}(s)$ on disjoint rollouts. \textbf{(c)} Held-out sequence RMS log-probability drift $M_{48}(s)$ (nats, log scale). \textbf{(d)} Pass-4 held-out directional efficiency across shared-\texttt{SymMax} token weighting schemes at nominal and reduced learning rates.}
\label{fig:app_multipass_bridge}
\vspace{-4pt}
\end{figure}

\paragraph{Pass-Wise Gradient Norms and Held-Out Alignment Across Reuse Passes.}
During online training, unclipped AsymRE exhibits growing actor gradient norms across passes ($1.67\times$ above BRAT on average across passes) and GRPO/Dr.GRPO experience a nearly $3\times$ rise in hard-clipped tokens, whereas BRAT's actor gradient norm decreased from the first reuse pass to the fourth (Figure~\ref{fig:brat_geometry}c--d). In the four-pass held-out alignment diagnostic on the prompt cache of Section \ref{sec:exp_b}  (Figure~\ref{fig:app_multipass_bridge}; Table~\ref{tab:app_multipass_bridge}), removing DS (\texttt{SymMax + Linear}) causes a $10\times$ higher held-out sequence drift than BRAT at Pass~4 and degrades held-out directional efficiency. Retaining DS bounds drift and gives BRAT the highest held-out efficiency at Passes~2 and~4 (and \texttt{Centered + DS} at Pass~3). Holding \texttt{SymMax} fixed at Pass~4 across token-level weighting schemes (Figure~\ref{fig:app_multipass_bridge}d), BRAT also outperforms unclamped $k_3$-KL, DAPO \texttt{Clip-Higher} \citep{yu2025dapo}, \texttt{TOPR} \citep{leroux2025topr}, and nominal- or multiplier-matched \texttt{SAPO} \citep{gao2025sapo}.

\begin{figure}[!t]
\centering
\includegraphics[width=0.92\textwidth]{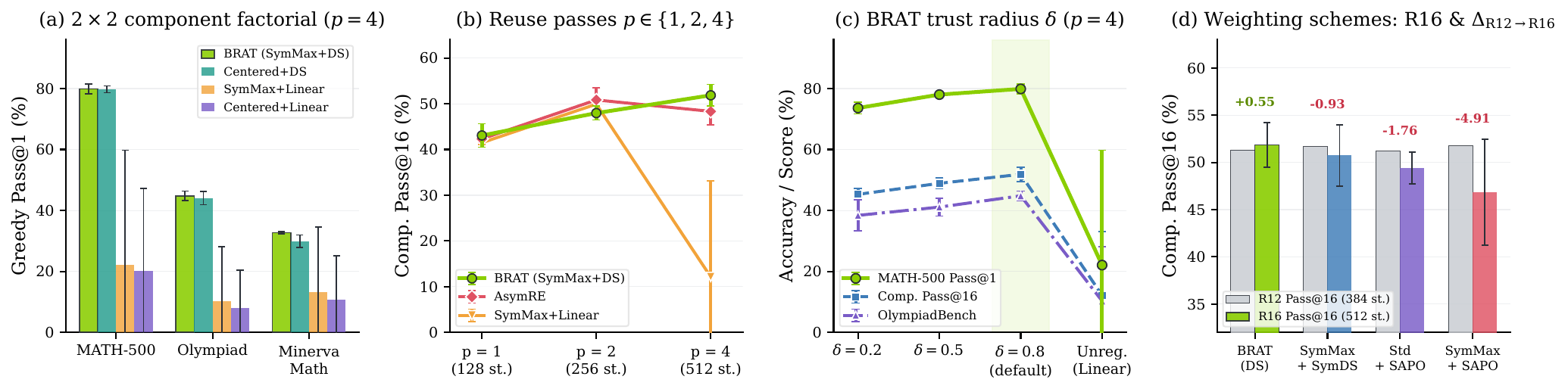}
\vspace{-5pt}
\caption{\textbf{Online component, reuse-depth, trust-radius, and token-weighting ablations in the 16-round 8-GPU suite on \texttt{Qwen3-8B-Base} ($\text{mean} \pm \text{SD}$ over 3 seeds; Tables~\ref{tab:app_online_ablations}--\ref{tab:app_token_gates}).} \textbf{(a)} $2\times 2$ component factorial ($p=4$, Round 16) on greedy MATH-500, OlympiadBench, and Minerva-Math. \textbf{(b)} Off-policy reuse passes $p \in \{1,2,4\}$ (Round 16) on 16-sample competition Pass@16. \textbf{(c)} Trust-region radius sensitivity ($p=4$, Round 16). \textbf{(d)} Token-weighting comparison (\textbf{BRAT} vs.\ \texttt{SymDS} and \texttt{SAPO}): competition Pass@16 retention under longer training.}
\label{fig:brat_online_ablations}
\vspace{-4pt}
\end{figure}

\paragraph{Component Factorial, Reuse Depth, and Trust Radius.}
Figure~\ref{fig:brat_online_ablations}a--c summarizes the 16-round online ablations in the 8-GPU suite (full Round-12 and Round-16 breakdowns in Appendix~\ref{app:brat_extended}, Table~\ref{tab:app_online_ablations}). In the $2\times 2$ factorial at $p=4$ (Figure~\ref{fig:brat_online_ablations}a), removing DS triggers multi-pass collapse under both group-centered scalings (\texttt{SymMax + Linear} and \texttt{Centered + Linear}); with DS, \texttt{Centered + DS} slightly leads on MATH-500 and Minerva-Math at Round~12, whereas by Round~16 \textbf{BRAT} (\texttt{SymMax + DS}) leads across all three greedy benchmarks and competition Avg@16 (with the largest gain on Minerva-Math). Across reuse depths $p \in \{1, 2, 4\}$ (Figure~\ref{fig:brat_online_ablations}b and Table~\ref{tab:app_online_ablations}b), \texttt{SymMax + Linear} improves from $p=1$ to $p=2$ before collapsing at $p=4$, while AsymRE avoids collapse via its negative baseline shift ($b_{\mathrm{asym}}=-0.05$) and attains higher single-sample MATH-500 and competition Avg@16 at $p=4$, but drops from $p=2$ to $p=4$ on both OlympiadBench Pass@1 and competition Pass@16. By contrast, BRAT improves monotonically across all five metrics from $p=1$ through $p=4$. Across trust radii $\delta \in \{0.2, 0.5, 0.8\}$ and the unregularized linear baseline at $p=4$ (Figure~\ref{fig:brat_online_ablations}c), $\delta=0.2$ underfits as predicted by~\eqref{eq:ds_impulse}, whereas $\delta=0.8$ achieves the highest accuracy and Pass@16 while removing DS collapses.

\paragraph{Longer-Horizon Token-Weighting Comparison and Pass@1 vs.\ Pass@16 Trade-Off.}
Extending training from Round~12 to Round~16 in the 8-GPU suite (Figure~\ref{fig:brat_online_ablations}d; full metrics in Appendix~\ref{app:brat_extended}, Table~\ref{tab:app_token_gates}) highlights two longer-horizon findings across token-level weighting schemes. First, \textbf{BRAT} (\texttt{SymMax + DS}) continues improving its mean across all five metrics through Round~16, outperforming the symmetric log-space variant (\texttt{SymDS}; Appendix~\ref{app:implementation}) on MATH-500, OlympiadBench, Minerva-Math, competition Avg@16, and competition Pass@16. Second, comparing token weighting schemes reveals a clear \emph{Pass@1 vs.\ Pass@16 trade-off}: replacing standard-deviation normalization in \texttt{Std + SAPO} with \texttt{SymMax} (\texttt{SymMax + SAPO}) reduces Round-12 variance across seeds on Minerva-Math, and \texttt{SymMax + SAPO} achieves higher greedy Pass@1 on OlympiadBench and Minerva-Math at Round~16 while losing $4.91$~pp in mean competition Pass@16 from Round~12 to Round~16. By contrast, BRAT maintained mean competition Pass@16 from Round~12 to Round~16 while the compared alternative schemes declined.

\section{Conclusion and Limitations}
\label{sec:conclusion}

We introduced \textbf{Bounded Residual-Aligned Training (BRAT)} to stabilize multi-pass rollout reuse in LLM post-training. Motivated by the nearly ninefold drop in directional efficiency from in-sample rollouts to disjoint held-out rollouts of the same prompts, BRAT combines a smooth Directional Saturating token objective with zero-sum, unit-bounded \texttt{SymMax} group normalization. Across multi-pass held-out alignment diagnostics, 16-round ablations, and six mathematical reasoning benchmarks, BRAT suppresses off-policy drift, prevents multi-pass collapse, maintained mean competition Pass@16 from Round~12 to Round~16, and improves accuracy while reducing variance across seeds.

\paragraph{Limitations and Future Work.}
Our study has several limitations. Theoretically, a gap between in-sample and held-out alignment remains inherent to small rollout groups, and Proposition~\ref{prop:proximal_identity}'s sequence-level guarantees do not transfer automatically to the token-level surrogate. Empirically, while our offline alignment diagnostic spans three base models, our online post-training experiments are conducted on a single base model (\texttt{Qwen3-8B-Base}) with LoRA adapters. Extending BRAT to additional model families, larger scales, full-parameter training, and hybrid schedules that combine Pass@16 retention with late-stage greedy Pass@1 refinement is an important direction for future work.

\clearpage
\bibliography{references}
\bibliographystyle{iclr2027_conference}

\newpage
\appendix

\section{Mathematical Proofs and Analytical Derivations}
\label{app:proofs}

\subsection{Exponential-Tilting Path and Variance Integral Identities}
\label{app:proof_dist}

Fix a prompt $x \in \X$ and suppress $x$ when unambiguous. For any intermediate inverse temperature $\tau \in [0,\eta]$, let $G_\tau(y) = \pi_0(y)e^{\tau R(y)}/Z_\tau$ denote the exponentially tilted distribution with log-partition function $\psi(\tau) \coloneqq \log Z_\tau$, mean reward $m_\tau \coloneqq \EE_{G_\tau}[R]$, and reward variance $V_\tau \coloneqq \VV_{G_\tau}(R)$. Let $r_{\max} \coloneqq \operatorname{ess\,sup}_{y\sim\pi_0} R(y)$ and $p_\xi \coloneqq \pi_0(\{y : R(y) \ge r_{\max} - \xi\})$ for reward tolerance $\xi > 0$.

\begin{proposition}[Exponential-Tilting Variance Integral Identities]
\label{prop:dist_flow}
For any bounded reward $R \in [0,1]$ and $\eta > 0$, the exponential-tilting family $\{G_\tau\}_{\tau\in[0,\eta]}$ satisfies $\partial_\tau G_\tau(y) = G_\tau(y)(R(y) - m_\tau)$ and the standard cumulant integral identities
\begin{equation}
\label{eq:app_variance_path}
m_\eta - m_0 = \int_0^\eta V_\tau\,d\tau,
\qquad
\KL(\pi_0 \Vert G_\eta) = \int_0^\eta (\eta - \tau)V_\tau\,d\tau,
\qquad
\KL(G_\eta \Vert \pi_0) = \int_0^\eta \tau\,V_\tau\,d\tau.
\end{equation}
Consequently, small-$\eta$ ($\eta \to 0$) and large-$\eta$ ($\eta > 0, p_\xi > 0$) regimes satisfy
\begin{align*}
m_\eta - m_0 &= \eta V_0 + O(\eta^2),
&\KL(\pi_0 \Vert G_\eta) &= \tfrac{1}{2}\eta^2 V_0 + O(\eta^3), \\
m_\eta - m_0 &\le r_{\max} - m_0,
&\KL(\pi_0 \Vert G_\eta) &\ge \eta(r_{\max} - \xi - m_0) + \log p_\xi.
\end{align*}
\end{proposition}

\begin{proof}
Since $R(y)\in[0,1]$ is bounded, $Z_\tau = \EE_{\pi_0}[e^{\tau R}]$ is smooth and strictly positive, with $\psi'(\tau) = m_\tau$ and $\psi''(\tau) = V_\tau \ge 0$. Differentiating $G_\tau(y) = \pi_0(y)\exp(\tau R(y) - \psi(\tau))$ gives $\partial_\tau G_\tau(y) = G_\tau(y)(R(y) - m_\tau)$, and integrating $m_\tau' = V_\tau$ over $[0,\eta]$ yields $m_\eta - m_0 = \int_0^\eta V_\tau\,d\tau$. Substituting $\log(G_\eta(y)/\pi_0(y)) = \eta R(y) - \psi(\eta)$ into the forward and reverse KL divergences gives
\begin{equation}
\label{eq:app_kl_cgf}
\KL(\pi_0 \Vert G_\eta) = \psi(\eta) - \psi(0) - \eta\psi'(0),
\qquad
\KL(G_\eta \Vert \pi_0) = \eta\psi'(\eta) - \bigl(\psi(\eta) - \psi(0)\bigr).
\end{equation}
Applying Taylor's theorem with integral remainder to $\psi(\eta)$ around $\tau=0$ yields the forward-KL integral in~\eqref{eq:app_variance_path}, and subtracting it from $\eta(m_\eta - m_0) = \eta\int_0^\eta V_\tau\,d\tau$ gives the reverse-KL integral. Finally, $m_\eta \le r_{\max}$ since $\mathrm{supp}(G_\eta) = \mathrm{supp}(\pi_0)$, and restricting $Z_\eta$ to $\{y: R(y)\ge r_{\max}-\xi\}$ gives $Z_\eta \ge p_\xi e^{\eta(r_{\max}-\xi)}$, which implies the lower bound on $\KL(\pi_0 \Vert G_\eta) = \psi(\eta) - \eta m_0$.
\end{proof}

\subsection{Proof of Proposition~\ref{prop:pullback} (Residual Decomposition of the Target Gradient)}
\label{app:proof_student}

\begin{proof}
Substituting $G_\eta(y\mid x) = \pi_0(y\mid x)(1 + a_\eta(x,y))$ from~\eqref{eq:signed_residual} into $\mathcal{L}_{G_\eta}(\theta)$ yields $\mathcal{L}_{G_\eta}(\theta) = \mathcal{L}_0(\theta) + \mathcal{L}_{\mathrm{RAT}}(\theta; \eta)$ by linearity of expectation. Differentiating under the expectation and using $\EE_{\pi_0}[a_\eta(x,Y)\mid x] = 0$ for $w_\eta \coloneqq 1 + a_\eta$ gives
\begin{equation*}
g_\eta(\theta) = -\nabla_\theta \mathcal{L}_{\mathrm{RAT}}(\theta; \eta)
= \EE_{\pi_0}\!\left[a_\eta(x,Y)\,s_\theta(x,Y)\right]
= \Cov_{\pi_0}\!\bigl(w_\eta(x,Y),\, s_\theta(x,Y)\bigr).
\end{equation*}
At $\pi_\theta = \pi_0$, the score identity $\EE_{\pi_0}[s_\theta(x,Y)\mid x] = 0$ implies $\nabla_\theta \mathcal{L}_0(\theta)\big|_{\pi_\theta=\pi_0} = 0$. For any update direction $d \in \R^{\dim(\theta)}$ with linearized shift $\delta_\theta(x,y; d) = s_\theta(x,y)^\top d$, we have $\langle g_\eta(\theta), d\rangle = \EE_{\pi_0}[a_\eta(x,Y)\delta_\theta(x,Y; d)]$; whenever $\EE_{\pi_0}[a_\eta^2] > 0$ and $\EE_{\pi_0}[\delta_\theta^2] > 0$, normalizing by $\sqrt{\EE_{\pi_0}[a_\eta^2]\,\EE_{\pi_0}[\delta_\theta^2]}$ gives~\eqref{eq:directional_efficiency}--\eqref{eq:rho_pred_def}, with $\rho_\eta^{\mathrm{pred}}(\theta; d) \in [-1,1]$ by Cauchy--Schwarz.
\end{proof}

\subsection{Proofs and Analytical Comparisons for BRAT (Propositions~\ref{prop:proximal_identity} and~\ref{prop:ds_gradient})}
\label{app:proof_brat}

\begin{proof}[Proof of Proposition~\ref{prop:proximal_identity}]
Fix $x\in\X$, suppress $x$ when unambiguous, and let $\mathcal{L}_t(p; x)$ denote the prompt-conditional integrand of~\eqref{eq:pop_brat_loss} so that $\mathcal{L}_t(p) = \EE_{x\sim\mathcal{D}}[\mathcal{L}_t(p; x)]$. Expanding $\KL(H_t \Vert p) - \KL(H_t \Vert q)$ for $H_t(y) = q(y)(1 + \delta_t A_q(y))$ gives
\begin{equation}
\KL(H_t \Vert p) - \KL(H_t \Vert q)
= \sum_{y\in\Y} q(y)\bigl(1 + \delta_t A_q(y)\bigr)\log\frac{q(y)}{p(y)}
= \delta_t \mathcal{L}_t(p; x),
\end{equation}
whose expectation over $x\sim\mathcal{D}$ proves Claim~1. For Claim~2, when $\mathrm{supp}(p(\cdot\mid x)) \subseteq \mathrm{supp}(q(\cdot\mid x))$, the density ratio $\rho_y \coloneqq p(y)/q(y)$ on $\mathrm{supp}(q)$ satisfies $\sum_{y\in\mathrm{supp}(q)} q(y)\rho_y = 1$ and $\alpha(x)(\EE_p[R\mid x] - \EE_q[R\mid x]) = \EE_{y\sim p}[A_q(y)]$. Since $\rho_y - 1 - \log \rho_y \ge 0$, $1 + A_q(y) \ge 0$, and $\EE_q[A_q] = 0$, we have for any $\delta_t \in (0, 1)$:
\begin{align}
\EE_{y\sim p}[A_q(y)] + \mathcal{L}_t(p; x)
&= \sum_{y} q(y)\!\left[A_q(y)\rho_y - A_q(y)\log \rho_y - \frac{1}{\delta_t}\log \rho_y\right] \nonumber\\
&= \sum_{y} q(y)\bigl(1 + A_q(y)\bigr)(\rho_y - 1 - \log \rho_y) + \left(\frac{1}{\delta_t} - 1\right)\KL(q \Vert p) \ge 0.
\end{align}
Taking expectations over $x\sim\mathcal{D}$ gives $J(p) - J(q) \ge -\mathcal{L}_t(p)$. For Claim~3, under $\TV(p,q) \coloneqq \frac{1}{2}\sum_y |p(y)-q(y)|$ and $R(x,y) \in [0,1]$, taking expectations over $x\sim\mathcal{D}$ at each round $t$ yields
\begin{align*}
\EE_{\mathcal{D}}\bigl[\EE_{H_t}[R] - \EE_{\pi_t}[R]\bigr] &= \delta_t \EE_{\mathcal{D}}\bigl[\EE_{\pi_t}[R A_{\pi_t}]\bigr] = \delta_t V(\pi_t), \\
\Bigl|\EE_{\mathcal{D}}\bigl[\EE_{\pi_{t+1}}[R] - \EE_{H_t}[R]\bigr]\Bigr| &\le \EE_{\mathcal{D}}\bigl[\TV(\pi_{t+1}, H_t)\bigr] = e_t;
\end{align*}
summing over $t=0,\dots,T-1$ yields~\eqref{eq:multi_round_bound}.
\end{proof}

\paragraph{Properties and Continuous-Reward Extension of \texttt{SymMax}.}
For any mixed binary group ($0 < \hat{p} < 1$), any non-negative linear scaling $A_i = \alpha(\hat{p})(R_i - \hat{p})$ satisfies $\frac{1}{K}\sum_{i=1}^K A_i = 0$ and $\max_i |A_i| = \alpha(\hat{p})\max(\hat{p}, 1-\hat{p})$; setting $\alpha(\hat{p}) = 1/\max(\hat{p}, 1-\hat{p})$ assigns unit-magnitude weights to minority outcomes ($|A_{\mathrm{minority}}^{\mathrm{BRAT}}| = 1$) while scaling centered rewards most strongly ($\alpha=2$) when the group contains equal numbers of correct and incorrect responses ($\hat{p}=0.5$). For general bounded continuous or multi-level rewards $R_i \in [r_{\min}, r_{\max}]$ with group mean $\hat{\mu} \coloneqq \frac{1}{K}\sum_{j=1}^K R_j$, \texttt{SymMax} generalizes directly via centered $\ell_\infty$ normalization:
\begin{equation*}
A_i^{\ell_\infty} \coloneqq \frac{R_i - \hat{\mu}}{\max_{1\le j\le K}|R_j - \hat{\mu}|}\,\mathbb{I}_{\max_j |R_j - \hat{\mu}| > 0} \in [-1, 1],
\end{equation*}
whose weights sum to zero within each group, assign unit magnitude to the furthest deviation from $\hat{\mu}$, and reduce identically to~\eqref{eq:symmax_def} when $R_i \in \{0,1\}$.

\begin{proof}[Proof of Proposition~\ref{prop:ds_gradient}]
When $A_i^{\mathrm{BRAT}} = 0$, $I_\delta(0) = [0,0]$ gives $c_{i,t} = 0$ and $\ell_{i,t}^{\mathrm{BRAT}}(\theta) \equiv 0$. For $A = A_i^{\mathrm{BRAT}} \ne 0$, inside the open interval $I_\delta(A)$, $c = z$ and $\psi_{\mathrm{DS}}(z, c) = e^z - 1 - z$, with derivative $\partial_z \psi_{\mathrm{DS}} = e^z - 1 = e^c - 1$. Outside $I_\delta(A)$, $c$ is constant at the boundary of $I_\delta(A)$ and $\psi_{\mathrm{DS}}(z, c)$ is affine in $z$ with slope $e^c - 1$. Because $z \mapsto e^{\operatorname{clamp}(z, I_\delta(A))} - 1$ is continuous across both endpoints of $I_\delta(A)$, $\psi_{\mathrm{DS}}$ is globally $C^1$ in $z$, yielding~\eqref{eq:brat_multiplier}--\eqref{eq:brat_multiplier_def}.
\end{proof}

\paragraph{Analytical Comparison with $k_3$-KL, SPO, and Integrated Gradient Multipliers.}
We summarize the three analytical comparisons referenced in Section~\ref{sec:ds_barrier}:
\begin{itemize}[leftmargin=1.6em,itemsep=2pt,topsep=2pt]
\item \textbf{Truncated $k_3$ Term vs.\ Unclamped KL Penalty:} Inside $I_\delta(A_i^{\mathrm{BRAT}})$, $\psi_{\mathrm{DS}}$ matches Schulman's $k_3$ KL estimator $e^z - 1 - z$ \citep{schulman2020kl}, whereas when a token log-ratio moves opposite to the advantage sign ($\operatorname{sign}(A_i^{\mathrm{BRAT}})z \le 0$) clamping gives $c_{i,t}=0$ and $\psi_{\mathrm{DS}}(z,0)=0$, so $\ell_{i,t}^{\mathrm{BRAT}}$ reduces to $-A_i^{\mathrm{BRAT}}z_{i,t}$ with unattenuated derivative $-A_i^{\mathrm{BRAT}}$ rather than an exponential penalty.
\item \textbf{Contrast with Simple Policy Optimization \citep[SPO;][]{xie2025spo}:} In terms of $\rho = e^z$, SPO uses a quadratic ratio penalty whose log-ratio derivative is
\begin{equation*}
\ell_{\mathrm{SPO}}(z; A, \epsilon) = -A\rho + \frac{|A|}{2\epsilon}(\rho - 1)^2,
\qquad
\frac{\partial \ell_{\mathrm{SPO}}}{\partial z} = -A\rho\left(1 - \operatorname{sign}(A)\frac{\rho - 1}{\epsilon}\right).
\end{equation*}
Unlike~\eqref{eq:brat_multiplier}, SPO retains the importance-ratio factor $\rho = e^z$, reverses gradient sign once $\operatorname{sign}(A)(\rho - 1) > \epsilon$, and grows quadratically in $\rho$ when $A<0$ and $\rho > 1$.
\item \textbf{Comparing Integrated Gradient Multipliers and Policy Staleness Within a Pass:} Over the positive-advantage interval $\rho = e^z \in [1, 1+\epsilon]$ for PPO and $\rho = e^z \in [1, 1+\delta]$ for BRAT, PPO's multiplier within the clipping interval is $m_{\mathrm{PPO}}(z) = e^z$ with integral $I_{\mathrm{PPO}}^+(\epsilon) = \int_0^{\ln(1+\epsilon)} e^z\,dz = \epsilon$, whereas integrating $m_{\mathrm{BRAT}}(z, +1) = 1 - (e^z - 1)/\delta$ yields~\eqref{eq:ds_impulse}. Under \texttt{SymMax}, minority-outcome trajectories satisfy $|A_i^{\mathrm{BRAT}}|=1$, matching the $1/\delta$ coefficient of~\eqref{eq:pop_brat_loss}, while majority-outcome trajectories share the sign-dependent boundary $1+\operatorname{sign}(A_i^{\mathrm{BRAT}})\delta$ with proportionally scaled step sizes. Because each round of $M=64$ prompts is updated in mini-batches of $\mathrm{mb}=8$ prompts ($8$ steps per pass), $\pi_\theta$ already shifts away from $q$ during steps $2\text{--}8$ of Pass~1.
\end{itemize}

\section{Training Hyperparameters and Implementation Details}
\label{app:implementation}

\paragraph{Numerical Precision and Symmetric Signed-Log Baseline.}
Table~\ref{tab:app_hyperparams} lists the hyperparameters for all online post-training runs on \texttt{Qwen3-8B-Base}. All token log-ratio arithmetic in~\eqref{eq:brat_token_loss} is computed in \texttt{float32}; for $|c_{i,t}| < 10^{-3}$, $e^c - 1 - c$ is evaluated via its fifth-order Taylor series to avoid floating-point cancellation. In the symmetric signed-log ablation (\texttt{SymDS}, Figure~\ref{fig:brat_online_ablations}d and Table~\ref{tab:app_token_gates}), let $v_{i,t} \coloneqq \operatorname{sign}(A_i^{\mathrm{BRAT}})z_{i,t}$; the directional clamp in~\eqref{eq:directional_bounds}--\eqref{eq:brat_token_loss} is replaced by a symmetric log-space bound applied to $v_{i,t}$:
\begin{align*}
c_{i,t}^{\mathrm{SymDS}} &\coloneqq \operatorname{clamp}\!\bigl(v_{i,t},\, [0, \ln(1+\delta)]\bigr),
\qquad
m_{\mathrm{SymDS}}\bigl(z_{i,t}, A_i^{\mathrm{BRAT}}\bigr) = 1 - \frac{e^{c_{i,t}^{\mathrm{SymDS}}}-1}{\delta} \in [0,1], \\
\ell_{i,t}^{\mathrm{SymDS}} &\coloneqq -A_i^{\mathrm{BRAT}}z_{i,t} + \frac{|A_i^{\mathrm{BRAT}}|}{\delta}\psi_{\mathrm{DS}}\bigl(v_{i,t}, c_{i,t}^{\mathrm{SymDS}}\bigr).
\end{align*}
For GRPO and Dr.GRPO, $c_{\mathrm{clip}} = 3.0$ denotes the standard dual-clip upper threshold on negative-advantage probability ratios.

\paragraph{Length Normalization, Decontamination, and Decoding Protocol.}
Following \citet{liu2025understanding}, token losses are summed over each response and divided by the constant length normalization $C = 4096$ (generation limit $8192$ tokens). Both the single-GPU main setting (via gradient accumulation) and the 8-GPU ablation suite (via data parallelism) use an identical global mini-batch of $\mathrm{mb}=8$ prompts ($64$ trajectories) per step. Training prompts from \texttt{NuminaMath-1.5} are decontaminated via exact match after string normalization against all six evaluation sets, and binary rewards $R \in \{0,1\}$ are verified by symbolic equivalence (\texttt{Math-Verify}). Evaluation uses deterministic decoding ($\text{temperature}=0$) for Greedy Pass@1 and $\text{temperature}=1.0, \text{top-}p=1.0$ (seed $20261027$, $8192$-token limit) for Pass@4 and 16-sample competition Avg@16 / Pass@16 (macro-averaged over AMC 2023, AIME 2024, and AIME 2025).

\begin{table}[!t]
\centering
\caption{Hyperparameters for multi-round online post-training on \texttt{Qwen3-8B-Base}.}
\label{tab:app_hyperparams}
\vspace{2pt}
\resizebox{\textwidth}{!}{%
\begin{tabular}{ll}
\toprule
\textbf{Parameter} & \textbf{Value} \\
\midrule
Base model & \texttt{Qwen3-8B-Base} (bf16) \\
LoRA configuration & $r=32, \alpha=32$, zero dropout, all attention \& MLP linear projections \\
Training corpus & \texttt{NuminaMath-1.5} verified competition math \\
Online rounds ($T$) & Main single-GPU setting: $12$ rounds ($384$ updates at $p=4$); 8-GPU ablation suite: $16$ rounds ($512$ updates, evaluated at $384$ and $512$) \\
Prompts per round ($M$) & $64$ ($768$ prompts at Round 12; $1024$ prompts at Round 16) \\
Rollouts per prompt ($K$) & $8$ (sampling temperature $1.0$, top-$p = 1.0$) \\
Max prompt / response length & $4096$ / $8192$ tokens \\
Length normalization scale ($C$) & $4096$ (\texttt{seq-mean-token-sum-norm}) \\
Mini-batch size ($\mathrm{mb}$) & $8$ prompts ($64$ trajectories per gradient step) \\
Off-policy reuse passes ($p$) & $4$ default ($32$ gradient steps per round; $p \in \{1,2,4\}$ in reuse sweep) \\
Optimizer \& learning rate & AdamW, peak LR $1.6 \times 10^{-5}$, 16-round cosine schedule ($\text{min\_lr\_ratio}=0.35$; 12-round runs evaluate the first 12 rounds of this schedule) \\
Method-specific parameters & \brat: $\delta=0.8$; AsymRE: $b_{\mathrm{asym}}=-0.05$; Dr.GRPO: $\epsilon=0.2, c_{\mathrm{clip}}=3.0$; GRPO: $\epsilon=0.2, c_{\mathrm{clip}}=3.0, \beta_{\mathrm{KL}}=0.01$; SAPO: $\tau_+=1.0, \tau_-=1.05$ \\
Random seeds & $3$ seeds per configuration (8-GPU ablation suite $\{20261021, 20261022, 20261024\}$; per-seed main-setting breakdown in Table~\ref{tab:app_per_seed}) \\
\bottomrule
\end{tabular}%
}
\end{table}

\section{Additional Verification for Residual Geometry (Experiments A and B)}
\label{app:exp_ab_details}

\begin{figure}[!t]
\centering
\includegraphics[width=0.86\textwidth]{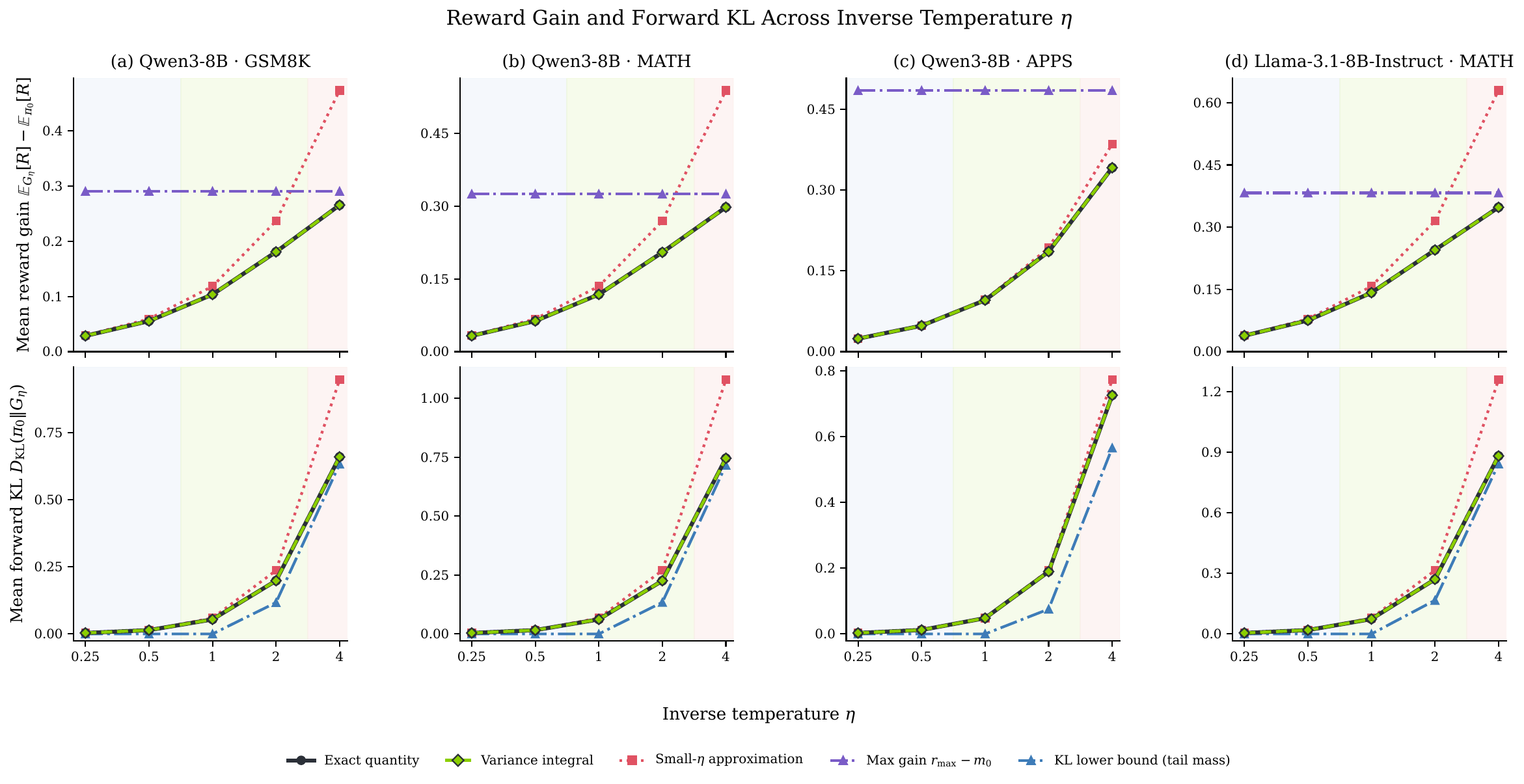}
\vspace{-2pt}
\caption{\textbf{Experiment A: Distributional residual regimes across four model--benchmark settings} ($128$ prompts, $K=48$ evaluation split, $\eta \in \{0.25, 0.5, 1, 2, 4\}$). \textbf{Top:} Mean reward gain $m_\eta - m_0$. \textbf{Bottom:} Mean forward KL $\KL(\pi_0 \Vert G_\eta)$.}
\label{fig:exp_a_main}
\end{figure}

\paragraph{Experiment A: Exponential-Tilting Variance Integral Verification.}
Figure~\ref{fig:exp_a_main} and Table~\ref{tab:app_exp_a} verify Proposition~\ref{prop:dist_flow} across four model--benchmark settings (\texttt{Qwen3-8B} on GSM8K, MATH, and APPS, plus \texttt{Llama-3.1-8B-Instruct} on MATH), averaging over prompts with non-zero empirical reward variance on the $K=48$ evaluation split ($22/128, 55/128, 63/128$, and $107/128$ prompts, respectively). Across all $20$ model--task--$\eta$ cells, numerical quadrature of the variance integrals in~\eqref{eq:app_variance_path} matches exact finite-support values to within $1.04\times 10^{-6}$, whereas first-order small-$\eta$ approximations depart substantially once $\eta \ge 2$.

\begin{figure}[!htbp]
\centering
\includegraphics[width=0.92\textwidth]{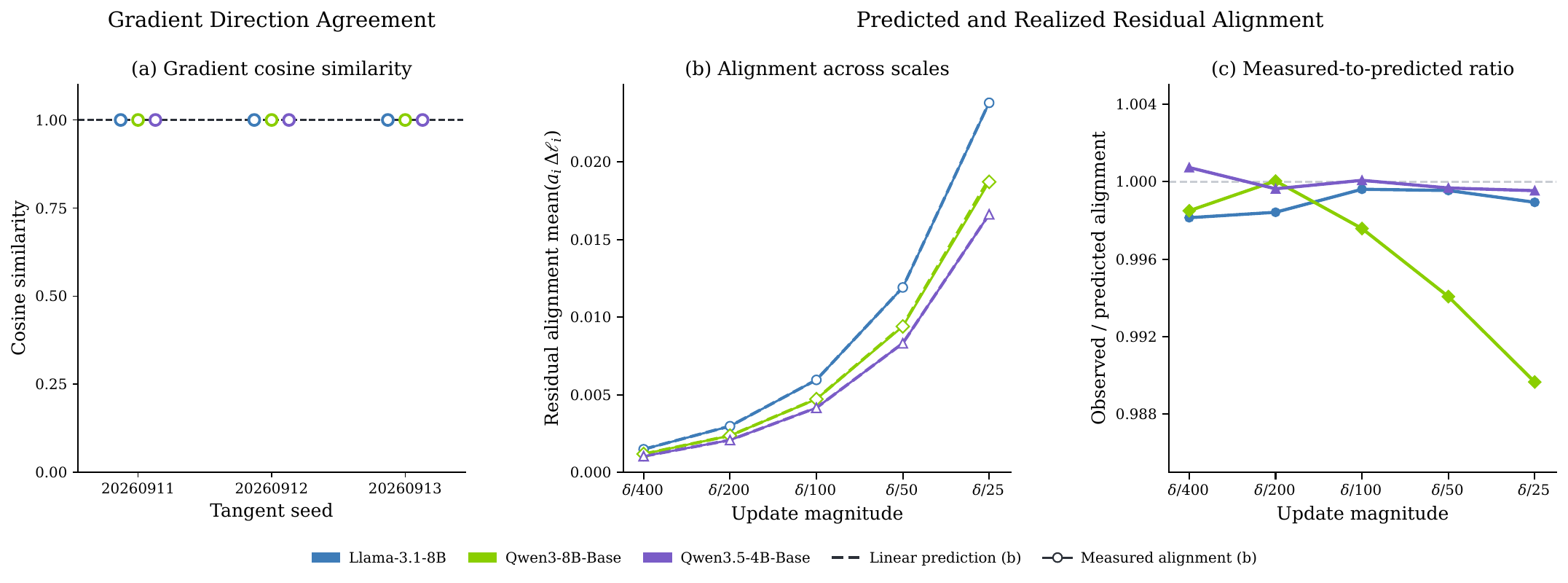}
\vspace{-2pt}
\caption{\textbf{Residual gradient identity and local finite-step alignment verification:} \textbf{(a)} Cosine agreement between the direct target-minus-reference gradient and the centered residual gradient ($\le 8.11\times 10^{-5}$ relative vector error). \textbf{(b, c)} Predicted vs.\ measured odd finite-step residual alignment and measured-to-predicted alignment ratio across five perturbation scales $\varepsilon$.}
\label{fig:app_b_identity}
\end{figure}

\begin{figure}[!htbp]
\centering
\includegraphics[width=0.92\textwidth]{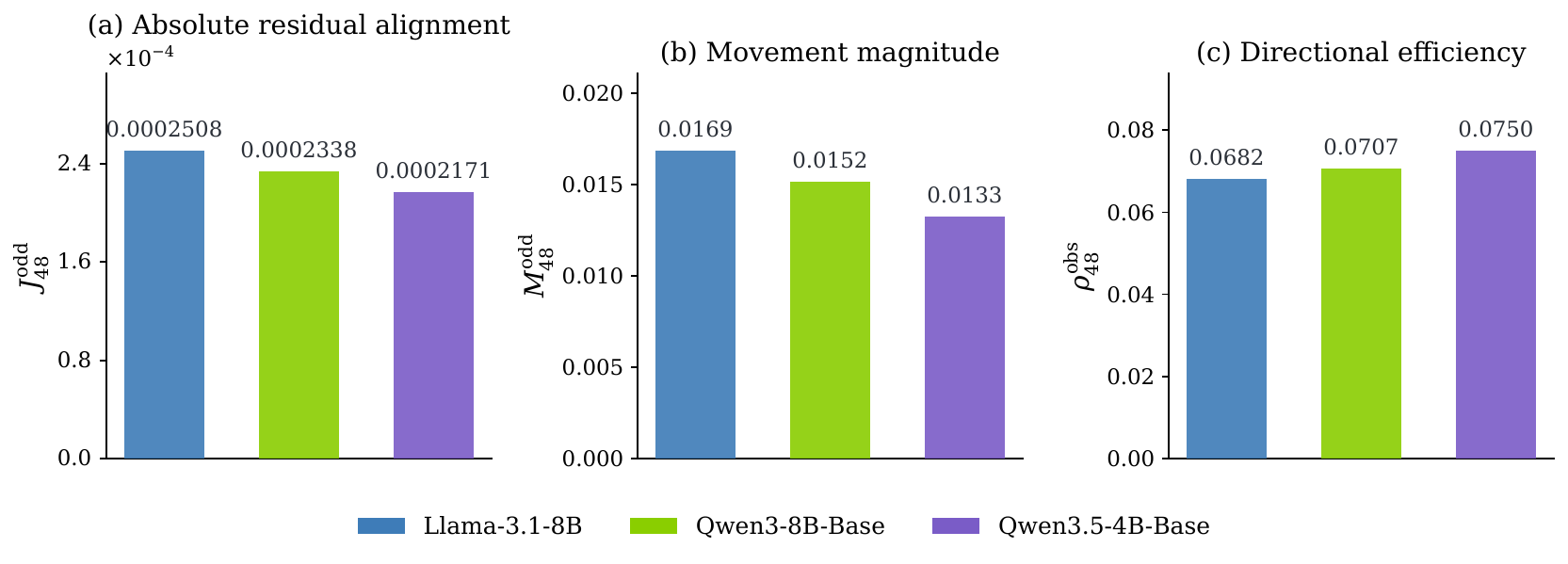}
\vspace{-2pt}
\caption{\textbf{Held-out alignment decomposition on the $K=48$ evaluation split ($\eta = 1.0$):} \textbf{(a)} Absolute odd residual alignment $J_{48}^{\mathrm{odd}}$. \textbf{(b)} RMS log-probability movement magnitude $M_{48}^{\mathrm{odd}}$. \textbf{(c)} Movement-normalized directional efficiency $\rho_{48}^{\mathrm{obs}}$ across all three base models.}
\label{fig:app_b_decomp}
\end{figure}

\paragraph{Experiment B: Preconditioned Step and Finite-Step Protocol.}
In Experiment~B (Figures~\ref{fig:app_b_identity}--\ref{fig:app_b_decomp} and Table~\ref{tab:app_exp_b}), $128$ MATH prompts ($64$ fixed rollouts per prompt sampled from \texttt{Qwen3-8B} so all three base models are evaluated on identical response sets) are partitioned into $K=16$ in-sample rollouts and $K=48$ disjoint held-out rollouts. On the in-sample split, each base model computes its empirical residual gradient $\hat{g}_{\eta,16}$ for~\eqref{eq:rat_grad} and damped diagonal-preconditioned direction $d = \hat{d}$ (using the empirical score second-moment diagonal $\operatorname{diag}\hat{F}_{16}$ with positive diagonal entries $\operatorname{diag}\hat{F}_{16}^+
\coloneqq \bigl((\hat{F}_{16})_{jj}\bigr)_{j:\,(\hat{F}_{16})_{jj}>0}$) and first-order gain $A_{\mathrm{diag}} = \langle \hat{g}_{\eta,16}, \hat{d}\rangle$. On any split $S \in \{16, 48\}$, we evaluate odd central-difference log-probability shifts $\delta^{\mathrm{odd}}(\varepsilon)$ and realized directional efficiency $\rho_{\eta,S}^{\mathrm{obs}}$:
\begin{align*}
\hat{d} &\coloneqq \bigl(\operatorname{diag}\hat{F}_{16} + \lambda I\bigr)^{-1}\hat{g}_{\eta,16},
\qquad
\lambda = 10^{-2}\operatorname{median}\bigl(\operatorname{diag}\hat{F}_{16}^+\bigr),
\qquad
A_{\mathrm{diag}} \coloneqq \hat{g}_{\eta,16}^\top \hat{d}, \\
\delta^{\mathrm{odd}}(\varepsilon) &\coloneqq \frac{1}{2}\Bigl[\log\pi_{\theta_0+\varepsilon\hat{d}}(y\mid x) - \log\pi_{\theta_0-\varepsilon\hat{d}}(y\mid x)\Bigr],
\qquad
\rho_{\eta,S}^{\mathrm{obs}} \coloneqq \frac{\widehat{\EE}_S\bigl[a_\eta\,\delta^{\mathrm{odd}}(\varepsilon)\bigr]}{\sqrt{\widehat{\EE}_S[a_\eta^2]\,\widehat{\EE}_S\bigl[\delta^{\mathrm{odd}}(\varepsilon)^2\bigr]}},
\end{align*}
where step scale $\varepsilon > 0$ is calibrated to match in-sample RMS movement across models ($\frac{1}{2}\widehat{\EE}_{16}[(\varepsilon s_{\theta_0}^\top\hat{d})^2] = 3\times 10^{-4}$), and $J_{48}^{\mathrm{odd}} \coloneqq \widehat{\EE}_{48}[a_\eta\,\delta^{\mathrm{odd}}(\varepsilon)]$ and $M_{48}^{\mathrm{odd}} \coloneqq \sqrt{\widehat{\EE}_{48}[\delta^{\mathrm{odd}}(\varepsilon)^2]}$ denote the held-out odd alignment and RMS movement in Figure~\ref{fig:app_b_decomp}. Table~\ref{tab:app_exp_b} also reports three equal-size ($K'=16$) held-out blocks and $95\%$ permutation null intervals from $2{,}000$ within-prompt reward permutations: across \texttt{Llama-3.1-8B}, \texttt{Qwen3-8B-Base}, and \texttt{Qwen3.5-4B-Base}, realized alignment exceeds the permutation null by $z = 16.7, 17.3, 18.4$ ($p < 5\times 10^{-4}$) on the $K=16$ in-sample split and by $z = 5.4, 6.3, 7.7$ ($p < 5\times 10^{-4}$, or $2.6\text{--}3.9\times$ the $95\%$ null upper bound) on the $K=48$ held-out split, while equal-size $K'=16$ held-out blocks yield $\bar{\rho}_{16'}^{\mathrm{obs}} = 0.0645, 0.0698, 0.0737$ (confirming that the $9\times$ drop from $\rho_{16}^{\mathrm{pred}}$ to $\rho_{48}^{\mathrm{obs}}$ reflects high-dimensional in-sample fitting on $K=16$ rather than support-size rescaling).

\paragraph{Multi-Pass Held-Out Alignment Diagnostic and Token-Weighting Controls.}
Table~\ref{tab:app_multipass_bridge} reports the four-pass held-out alignment diagnostic (seed \texttt{20260911}, $16$ mini-batch steps per pass across the $128$ prompts at $\text{lr}=1.6\times 10^{-5}$ and $C=4096$) corresponding to Figure~\ref{fig:app_multipass_bridge}. Whereas Table~\ref{tab:app_exp_b} evaluates the single-step update at $\eta=1.0$ (held-out residual RMS $0.2180$), Table~\ref{tab:app_multipass_bridge} evaluates multi-pass updates against the $\eta=0.8$ target residual (held-out residual RMS $0.1771$) with BRAT trust radius $\delta=0.8$. In Table~\ref{tab:app_multipass_bridge}, the last three columns record the percentage of active tokens whose log-ratio shift $z_{i,t}$ falls into the opposite-sign region ($\operatorname{sign}(A)z \le 0$), the interior interval $I_{0.8}(A)$, and the saturated region ($\rho = e^z$ beyond $1\pm 0.8$): (i) for BRAT, `Sat.\ (\%)' displays `0.00' because only $0.0017\%$ of active tokens exceed $\rho = 1\pm 0.8$ at Pass~4, as linear attenuation within $I_{0.8}(A)$ damps shifts before they reach the boundary; and (ii) for AsymRE, subtracting a negative baseline shift ($A_i = R_i - \hat{p} + 0.05$) adds a positive likelihood term to every sampled response, shifting both correct ($+81.1$~nats) and incorrect ($+134.5$~nats) responses positively at Pass~4 and inflating the uncentered RMS movement denominator $M_{48}=98.41$~nats relative to the zero-sum residual numerator. On the $49$ mixed prompts at Pass~4, the paired prompt-bootstrap difference between BRAT and \texttt{Centered + DS} spans zero ($\Delta\rho_{48}=+0.0010$, $95\%$ CI $[-0.0137, +0.0172]$). In the bottom block of Table~\ref{tab:app_multipass_bridge}, holding \texttt{SymMax} fixed at Pass~4 shows that BRAT achieves the highest held-out efficiency ($\rho_{48}=0.04879$) across all controls and the lowest drift ($M_{48}=2.67$~nats) among nominal-LR ($\text{lr}=1.6\times 10^{-5}$) token weighting schemes---outperforming unclamped $k_3$-KL, DAPO \texttt{Clip-Higher}, \texttt{TOPR}, and \texttt{SAPO}---whereas the multiplier-matched \texttt{SAPO} ($6.5\times 10^{-6}$) and $1/4\times$-LR \texttt{Linear} ($4.0\times 10^{-6}$) controls yield smaller movement ($M_{48}=1.86\text{--}1.89$~nats) only with much lower held-out directional efficiency ($\rho_{48}=0.01802$ and $0.00047$).

\begin{table}[!htbp]
\centering
\caption{\textbf{Experiment A numeric summary on the $K=48$ evaluation split (corresponding to Figure~\ref{fig:exp_a_main}).}}
\label{tab:app_exp_a}
\vspace{2pt}
\resizebox{\textwidth}{!}{%
\begin{tabular}{llccccccc}
\toprule
\textbf{Model $\cdot$ Task} & $\boldsymbol{\eta}$ & \textbf{Exact Gain} & \textbf{Integral Gain} & \textbf{First-Order $\eta V_0$} & \textbf{Max Gain $r_{\max}-m_0$} & \textbf{Exact $\KL(\pi_0\Vert G_\eta)$} & \textbf{Integral KL} & \textbf{First-Order KL} \\
\midrule
\multirow{3}{*}{\texttt{Qwen3-8B} $\cdot$ GSM8K}
 & 0.5 & 0.0554 & 0.0554 & 0.0592 & 0.2907 & 0.0142 & 0.0142 & 0.0148 \\
 & 1.0 & 0.1036 & 0.1036 & 0.1184 & 0.2907 & 0.0542 & 0.0542 & 0.0592 \\
 & 4.0 & 0.2657 & 0.2657 & 0.4736 & 0.2907 & 0.6597 & 0.6597 & 0.9471 \\
\midrule
\multirow{3}{*}{\texttt{Qwen3-8B} $\cdot$ MATH}
 & 0.5 & 0.0632 & 0.0632 & 0.0674 & 0.3254 & 0.0162 & 0.0162 & 0.0169 \\
 & 1.0 & 0.1181 & 0.1181 & 0.1348 & 0.3254 & 0.0618 & 0.0618 & 0.0674 \\
 & 4.0 & 0.2981 & 0.2981 & 0.5393 & 0.3254 & 0.7455 & 0.7455 & 1.0786 \\
\midrule
\multirow{3}{*}{\texttt{Qwen3-8B} $\cdot$ APPS}
 & 0.5 & 0.0482 & 0.0482 & 0.0482 & 0.4853 & 0.0121 & 0.0121 & 0.0121 \\
 & 1.0 & 0.0956 & 0.0956 & 0.0965 & 0.4853 & 0.0481 & 0.0481 & 0.0482 \\
 & 4.0 & 0.3418 & 0.3418 & 0.3860 & 0.4853 & 0.7250 & 0.7250 & 0.7720 \\
\midrule
\multirow{3}{*}{\texttt{Llama-3.1-8B-Inst} $\cdot$ MATH}
 & 0.5 & 0.0753 & 0.0753 & 0.0787 & 0.3826 & 0.0191 & 0.0191 & 0.0197 \\
 & 1.0 & 0.1420 & 0.1420 & 0.1574 & 0.3826 & 0.0739 & 0.0739 & 0.0787 \\
 & 4.0 & 0.3479 & 0.3479 & 0.6298 & 0.3826 & 0.8821 & 0.8821 & 1.2596 \\
\bottomrule
\end{tabular}%
}
\end{table}

\begin{table}[!htbp]
\centering
\caption{\textbf{Experiment B numeric summary and equal-size / permutation controls across base models ($\eta = 1.0$, $128$ prompts).} Columns report decomposition error, preconditioned in-sample gain $A_{\mathrm{diag}}$, in-sample efficiency $\rho_{16}^{\mathrm{pred}}$, held-out efficiency $\rho_{48}^{\mathrm{obs}}$, mean efficiency $\bar{\rho}_{16'}^{\mathrm{obs}}$ across three disjoint $K'=16$ held-out blocks, and $95\%$ permutation null intervals.}
\label{tab:app_exp_b}
\vspace{2pt}
\resizebox{\textwidth}{!}{%
\begin{tabular}{lccccccc}
\toprule
\textbf{Base Model} & \textbf{Decomp.\ Err.} & $\boldsymbol{A_{\mathrm{diag}}}$ & $\boldsymbol{\rho_{16}^{\mathrm{pred}}}$ & $\boldsymbol{\rho_{48}^{\mathrm{obs}}}$ & \textbf{Equal-Size $\boldsymbol{\bar{\rho}_{16'}^{\mathrm{obs}}}$ (3 blocks)} & \textbf{$\boldsymbol{K{=}16}$ Perm.\ Null 95\% CI} & \textbf{$\boldsymbol{K{=}48}$ Perm.\ Null 95\% CI} \\
\midrule
\texttt{Llama-3.1-8B} & $2.94\times 10^{-5}$ & $1721.1$ & $0.5865$ & $0.0682$ & $0.0645$ ($0.0325 / 0.0987 / 0.0625$) & $[-0.0625, +0.0737]$ & $[-0.0227, +0.0259]$ \\
\texttt{Qwen3-8B-Base} & $8.11\times 10^{-5}$ & $1502.6$ & $0.6571$ & $0.0707$ & $0.0698$ ($0.0611 / 0.0848 / 0.0635$) & $[-0.0688, +0.0715]$ & $[-0.0227, +0.0215]$ \\
\texttt{Qwen3.5-4B-Base} & $3.89\times 10^{-5}$ & $832.5$ & $0.7148$ & $0.0750$ & $0.0737$ ($0.0615 / 0.0807 / 0.0789$) & $[-0.0691, +0.0759]$ & $[-0.0194, +0.0194]$ \\
\bottomrule
\end{tabular}%
}
\end{table}

\begin{table}[!htbp]
\centering
\caption{\textbf{Multi-pass $K=16 \to K=48$ held-out alignment diagnostics and Pass-4 token-weighting controls on \texttt{Qwen3-8B-Base} ($128$ shared prompts, corresponding to Figure~\ref{fig:app_multipass_bridge}).} Columns report in-sample and held-out directional efficiency ($\rho_{16}, \rho_{48}$), sequence RMS log-probability drift ($M_{16}, M_{48}$ in nats), token log-ratio quantiles ($|z|_{p95}, |z|_{\max}$), and token occupancy (\%) across the opposite-sign (\textbf{Lin.}), interior (\textbf{Atten.}), and saturated (\textbf{Sat.}) log-ratio regions of $I_{0.8}(A)$.}
\label{tab:app_multipass_bridge}
\vspace{2pt}
\setlength{\tabcolsep}{4.0pt}
\resizebox{\textwidth}{!}{%
\begin{tabular}{llccccccccc}
\toprule
\textbf{Method} & \textbf{Pass / Setting} & $\boldsymbol{\rho_{16}\;(\uparrow)}$ & $\boldsymbol{M_{16}}$ & $\boldsymbol{\rho_{48}\;(\uparrow)}$ & $\boldsymbol{M_{48}\;(\downarrow\text{drift})}$ & $\boldsymbol{|z|_{p95}}$ & $\boldsymbol{|z|_{\max}}$ & \textbf{Lin.\ (\%)} & \textbf{Atten.\ (\%)} & \textbf{Sat.\ (\%)} \\
\midrule
\multirow{4}{*}{\brat\textbf{ (\texttt{SymMax + DS})}}
 & Pass 1 & 0.1256 & 1.37 & 0.00486 & 1.30 & 0.0776 & \textbf{0.508} & 53.59 & 46.41 & 0.00 \\
 & Pass 2 & 0.2976 & 1.84 & \textbf{0.01649} & 1.70 & 0.0849 & 0.737 & 52.84 & 47.16 & 0.00 \\
 & Pass 3 & \textbf{0.5093} & 2.38 & 0.04205 & 1.86 & 0.0901 & 0.859 & 53.14 & 46.86 & 0.00 \\
 & Pass 4 & 0.5638 & 3.76 & \textbf{0.04879} & 2.67 & 0.1010 & 1.122 & 53.61 & 46.39 & 0.00 \\
\midrule
\multirow{4}{*}{\texttt{Centered + DS} (Ablation)}
 & Pass 1 & \textbf{0.1372} & \textbf{1.28} & 0.00467 & \textbf{1.20} & 0.0783 & 0.665 & 53.76 & 46.24 & 0.00 \\
 & Pass 2 & \textbf{0.3198} & \textbf{1.81} & 0.01278 & \textbf{1.67} & 0.0854 & 0.866 & 52.86 & 47.14 & 0.00 \\
 & Pass 3 & 0.4869 & \textbf{2.27} & \textbf{0.04440} & \textbf{1.80} & 0.0898 & 0.748 & 53.08 & 46.92 & 0.00 \\
 & Pass 4 & \textbf{0.5742} & \textbf{3.56} & 0.04759 & \textbf{2.51} & 0.0989 & 1.234 & 52.98 & 47.02 & 0.00 \\
\midrule
\multirow{4}{*}{\texttt{SymMax + Linear} (Ablation)}
 & Pass 1 & 0.0617 & 3.79 & 0.00948 & 3.68 & 0.0876 & 0.795 & 52.40 & 47.60 & 0.00 \\
 & Pass 2 & 0.1058 & 6.71 & 0.01210 & 6.60 & 0.1140 & 1.374 & 48.91 & 51.09 & 0.00 \\
 & Pass 3 & 0.1115 & 13.42 & 0.01385 & 13.37 & 0.1410 & 2.746 & 48.67 & 51.32 & 0.00 \\
 & Pass 4 & 0.1175 & 27.45 & 0.01874 & 27.21 & 0.1911 & 4.798 & 48.49 & 51.46 & 0.05 \\
\midrule
\multirow{4}{*}{Dr.GRPO}
 & Pass 1 & 0.1171 & 1.39 & $-0.00280$ & 1.35 & 0.0794 & 0.623 & 53.41 & 46.59 & 0.00 \\
 & Pass 2 & 0.2133 & 2.42 & 0.01199 & 2.34 & 0.0862 & 0.705 & 51.87 & 48.13 & 0.00 \\
 & Pass 3 & 0.3581 & 2.72 & 0.03595 & 2.43 & 0.0880 & 0.742 & 52.24 & 47.76 & 0.00 \\
 & Pass 4 & 0.4760 & 3.46 & 0.04308 & 2.79 & 0.0929 & 0.984 & 52.74 & 47.26 & 0.00 \\
\midrule
\multirow{4}{*}{GRPO}
 & Pass 1 & 0.1326 & 1.52 & \textbf{0.01602} & 1.44 & 0.0795 & 1.227 & 53.62 & 46.38 & 0.00 \\
 & Pass 2 & 0.2437 & 2.35 & 0.01635 & 2.15 & 0.0850 & 0.734 & 51.70 & 48.30 & 0.00 \\
 & Pass 3 & 0.3747 & 2.99 & 0.02437 & 2.55 & 0.0904 & 0.997 & 51.61 & 48.39 & 0.00 \\
 & Pass 4 & 0.4965 & 3.78 & 0.04326 & 2.92 & 0.0977 & 1.109 & 52.40 & 47.60 & 0.00 \\
\midrule
\multirow{4}{*}{AsymRE}
 & Pass 1 & 0.0115 & 30.86 & 0.00053 & 30.50 & 0.1524 & 5.753 & 65.78 & 33.62 & 0.60 \\
 & Pass 2 & 0.0156 & 52.80 & 0.00025 & 52.20 & 0.4784 & 11.013 & 65.61 & 31.92 & 2.48 \\
 & Pass 3 & 0.0187 & 76.93 & 0.00006 & 76.22 & 0.5348 & 16.358 & 64.58 & 32.38 & 3.04 \\
 & Pass 4 & 0.0253 & 99.24 & 0.00074 & 98.41 & 0.6182 & 18.033 & 62.13 & 34.36 & 3.51 \\
\midrule
\midrule
\multicolumn{11}{l}{\textbf{Token-Weighting \& Step-Size Controls at Pass 4 (Shared \texttt{SymMax} Residual)}} \\
\midrule
\brat\textbf{ (\texttt{SymMax + DS}, $\delta=0.8$)} & Pass 4 ($\text{lr}=1.6\times 10^{-5}$) & $\mathbf{0.5638}$ & 3.76 & $\mathbf{0.04879}$ & $\mathbf{2.67}$ & 0.1010 & 1.122 & 53.61 & 46.39 & 0.00 \\
\texttt{SymMax + Unclamped $k_3$} ($\delta=0.8$) & Pass 4 ($\text{lr}=1.6\times 10^{-5}$) & 0.5114 & 3.78 & 0.04495 & 2.98 & 0.0959 & 1.234 & 53.28 & 46.72 & 0.00 \\
\texttt{SymMax + Clip-Higher} ($\epsilon_-=0.2, \epsilon_+=0.28$, DAPO) & Pass 4 ($\text{lr}=1.6\times 10^{-5}$) & 0.5516 & 3.52 & 0.03630 & 2.73 & 0.1020 & 1.109 & 53.53 & 46.47 & 0.00 \\
\texttt{SymMax + TOPR} (Tapered REINFORCE) & Pass 4 ($\text{lr}=1.6\times 10^{-5}$) & 0.2668 & 9.06 & 0.01812 & 8.49 & 0.1233 & 1.746 & 54.61 & 45.31 & 0.07 \\
\texttt{SymMax + SAPO} ($\tau_+=1.0, \tau_-=1.05$) & Pass 4 ($\text{lr}=1.6\times 10^{-5}$) & 0.1842 & 12.11 & 0.01727 & 11.74 & 0.1210 & 1.969 & 54.37 & 45.57 & 0.06 \\
\cmidrule(lr){1-11}
\texttt{SymMax + SAPO} (multiplier-matched LR) & Pass 4 ($\text{lr}=6.5\times 10^{-6}$) & 0.2418 & 1.98 & 0.01802 & 1.89 & 0.0922 & 0.795 & 50.84 & 49.16 & 0.00 \\
\texttt{SymMax + Linear} ($1/4\times$ LR control) & Pass 4 ($\text{lr}=4.0\times 10^{-6}$) & 0.1081 & 1.88 & 0.00047 & 1.86 & 0.0873 & 0.602 & 51.12 & 48.88 & 0.00 \\
\bottomrule
\end{tabular}%
}
\end{table}

\section{Extended Online Ablations, Per-Seed Breakdown, and Runtime Analysis}
\label{app:brat_extended}

\begin{table}[!t]
\centering
\caption{\textbf{Component, reuse-budget, and trust-radius ablations across Round 12 and Round 16 on \texttt{Qwen3-8B-Base}} ($\text{mean} \pm \text{SD}$ in \% across 3 seeds in the 16-round 8-GPU ablation suite; Round 16 corresponds to Figure~\ref{fig:brat_online_ablations}a--c).}
\label{tab:app_online_ablations}
\vspace{2pt}
\setlength{\tabcolsep}{3.0pt}
\resizebox{\textwidth}{!}{%
\begin{tabular}{llcccccccc}
\toprule
\textbf{Ablation Block} & \textbf{Configuration} & \textbf{Horizon} & $\boldsymbol{p}$ & \textbf{Steps} & \textbf{MATH-500} & \textbf{OlympiadBench} & \textbf{Minerva-Math} & \textbf{Comp.\ Avg@16} & \textbf{Comp.\ Pass@16} \\
\midrule
\multirow{4}{*}{\textbf{(a) $2\times 2$ Factorial}}
 & \brat\textbf{ (\texttt{SymMax + DS}, $\delta=0.8$)} & R12 / R16 & 4 & 384 / 512 & $79.27 \pm 1.03$ / $\mathbf{79.87 \pm 1.62}$ & $\mathbf{43.62 \pm 0.39}$ / $\mathbf{44.81 \pm 1.51}$ & $29.41 \pm 0.64$ / $\mathbf{32.72 \pm 0.37}$ & $\mathbf{23.32 \pm 0.88}$ / $\mathbf{25.08 \pm 0.39}$ & $\mathbf{51.30 \pm 0.98}$ / $51.85 \pm 2.36$ \\
 & \texttt{Centered + DS} ($\delta=0.8$) & R12 / R16 & 4 & 384 / 512 & $\mathbf{80.00 \pm 0.72}$ / $79.73 \pm 1.14$ & $\mathbf{43.62 \pm 1.21}$ / $44.02 \pm 2.15$ & $\mathbf{30.15 \pm 1.33}$ / $29.90 \pm 2.09$ & $22.91 \pm 1.66$ / $24.03 \pm 0.82$ & $49.63 \pm 2.78$ / $\mathbf{52.04 \pm 3.31}$ \\
 & \texttt{Centered + Linear} (Unregularized) & R12 / R16 & 4 & 384 / 512 & $58.80 \pm 26.58$ / $20.13 \pm 27.04$ & $27.94 \pm 19.38$ / $7.96 \pm 12.51$ & $25.49 \pm 12.76$ / $10.66 \pm 14.42$ & $16.68 \pm 12.61$ / $4.46 \pm 7.40$ & $37.78 \pm 19.03$ / $15.93 \pm 23.78$ \\
 & \texttt{SymMax + Linear} (Unregularized) & R12 / R16 & 4 & 384 / 512 & $23.13 \pm 39.72$ / $22.07 \pm 37.70$ & $11.23 \pm 19.44$ / $10.29 \pm 17.82$ & $12.25 \pm 20.28$ / $13.24 \pm 21.34$ & $5.18 \pm 8.93$ / $4.73 \pm 8.20$ & $10.65 \pm 17.73$ / $12.13 \pm 21.01$ \\
\midrule
\multirow{9}{*}{\textbf{(b) Reuse Passes $p$}}
 & \brat\textbf{ (\texttt{SymMax + DS}, $p=1$)} & R12 / R16 & 1 & 96 / 128 & $\mathbf{70.60 \pm 0.87}$ / $72.40 \pm 1.71$ & $\mathbf{36.15 \pm 1.94}$ / $\mathbf{37.64 \pm 2.44}$ & $\mathbf{23.65 \pm 0.56}$ / $\mathbf{25.25 \pm 1.39}$ & $\mathbf{13.07 \pm 0.87}$ / $\mathbf{13.83 \pm 0.27}$ & $42.13 \pm 3.05$ / $\mathbf{43.06 \pm 2.55}$ \\
 & \texttt{SymMax + Linear} ($p=1$) & R12 / R16 & 1 & 96 / 128 & $69.60 \pm 0.53$ / $\mathbf{72.80 \pm 0.87}$ & $32.74 \pm 0.52$ / $35.46 \pm 1.32$ & $23.16 \pm 0.74$ / $23.65 \pm 1.06$ & $12.52 \pm 1.17$ / $13.28 \pm 0.71$ & $\mathbf{42.69 \pm 2.80}$ / $41.48 \pm 0.42$ \\
 & AsymRE ($p=1$) & R12 / R16 & 1 & 96 / 128 & $69.20 \pm 0.92$ / $71.80 \pm 0.92$ & $33.38 \pm 1.04$ / $35.81 \pm 1.98$ & $23.41 \pm 0.56$ / $23.41 \pm 1.18$ & $12.17 \pm 0.23$ / $12.79 \pm 0.71$ & $39.44 \pm 3.48$ / $42.31 \pm 0.70$ \\
\cmidrule(lr){2-10}
 & \brat\textbf{ (\texttt{SymMax + DS}, $p=2$)} & R12 / R16 & 2 & 192 / 256 & $78.20 \pm 0.53$ / $78.60 \pm 1.51$ & $42.28 \pm 1.03$ / $43.92 \pm 0.77$ & $28.19 \pm 1.53$ / $29.29 \pm 0.21$ & $17.07 \pm 1.77$ / $18.63 \pm 1.32$ & $\mathbf{51.30 \pm 0.89}$ / $47.96 \pm 1.53$ \\
 & \texttt{SymMax + Linear} ($p=2$) & R12 / R16 & 2 & 192 / 256 & $\mathbf{79.47 \pm 1.14}$ / $79.13 \pm 0.83$ & $\mathbf{44.36 \pm 0.53}$ / $\mathbf{44.41 \pm 0.56}$ & $\mathbf{30.27 \pm 1.49}$ / $\mathbf{31.00 \pm 1.53}$ & $\mathbf{20.31 \pm 1.33}$ / $\mathbf{20.71 \pm 0.63}$ & $50.74 \pm 1.31$ / $49.91 \pm 1.31$ \\
 & AsymRE ($p=2$) & R12 / R16 & 2 & 192 / 256 & $77.47 \pm 0.12$ / $\mathbf{80.00 \pm 0.00}$ & $44.16 \pm 1.49$ / $44.31 \pm 0.17$ & $27.82 \pm 1.39$ / $28.06 \pm 0.21$ & $18.84 \pm 1.22$ / $20.21 \pm 0.76$ & $50.09 \pm 1.95$ / $\mathbf{50.83 \pm 2.65}$ \\
\cmidrule(lr){2-10}
 & \brat\textbf{ (\texttt{SymMax + DS}, $p=4$)} & R12 / R16 & 4 & 384 / 512 & $79.27 \pm 1.03$ / $79.87 \pm 1.62$ & $\mathbf{43.62 \pm 0.39}$ / $\mathbf{44.81 \pm 1.51}$ & $\mathbf{29.41 \pm 0.64}$ / $\mathbf{32.72 \pm 0.37}$ & $23.32 \pm 0.88$ / $25.08 \pm 0.39$ & $\mathbf{51.30 \pm 0.98}$ / $\mathbf{51.85 \pm 2.36}$ \\
 & \texttt{SymMax + Linear} ($p=4$) & R12 / R16 & 4 & 384 / 512 & $23.13 \pm 39.72$ / $22.07 \pm 37.70$ & $11.23 \pm 19.44$ / $10.29 \pm 17.82$ & $12.25 \pm 20.28$ / $13.24 \pm 21.34$ & $5.18 \pm 8.93$ / $4.73 \pm 8.20$ & $10.65 \pm 17.73$ / $12.13 \pm 21.01$ \\
 & AsymRE ($p=4$) & R12 / R16 & 4 & 384 / 512 & $\mathbf{79.93 \pm 2.25}$ / $\mathbf{81.07 \pm 1.21}$ & $43.37 \pm 4.12$ / $42.43 \pm 1.82$ & $28.55 \pm 2.58$ / $28.19 \pm 5.32$ & $\mathbf{27.05 \pm 2.02}$ / $\mathbf{27.32 \pm 1.14}$ & $49.72 \pm 2.27$ / $48.33 \pm 2.90$ \\
\midrule
\multirow{4}{*}{\textbf{(c) Trust Radius $\delta$}}
 & \brat\textbf{ (\texttt{SymMax + DS}, $\delta=0.2$)} & R12 / R16 & 4 & 384 / 512 & $71.33 \pm 2.40$ / $73.60 \pm 1.91$ & $36.05 \pm 2.78$ / $38.38 \pm 5.08$ & $24.26 \pm 2.57$ / $26.59 \pm 1.85$ & $12.55 \pm 0.78$ / $14.21 \pm 0.56$ & $42.69 \pm 2.97$ / $45.28 \pm 1.94$ \\
 & \brat\textbf{ (\texttt{SymMax + DS}, $\delta=0.5$)} & R12 / R16 & 4 & 384 / 512 & $76.80 \pm 1.04$ / $78.00 \pm 0.00$ & $40.45 \pm 3.29$ / $41.10 \pm 2.92$ & $28.19 \pm 1.29$ / $28.92 \pm 1.74$ & $19.27 \pm 0.71$ / $20.86 \pm 0.36$ & $49.81 \pm 3.95$ / $48.89 \pm 1.73$ \\
 & \brat\textbf{ (\texttt{SymMax + DS}, $\delta=0.8$)} & R12 / R16 & 4 & 384 / 512 & $\mathbf{79.27 \pm 1.03}$ / $\mathbf{79.87 \pm 1.62}$ & $\mathbf{43.62 \pm 0.39}$ / $\mathbf{44.81 \pm 1.51}$ & $\mathbf{29.41 \pm 0.64}$ / $\mathbf{32.72 \pm 0.37}$ & $\mathbf{23.32 \pm 0.88}$ / $\mathbf{25.08 \pm 0.39}$ & $\mathbf{51.30 \pm 0.98}$ / $\mathbf{51.85 \pm 2.36}$ \\
 & \texttt{SymMax + Linear} (Unregularized) & R12 / R16 & 4 & 384 / 512 & $23.13 \pm 39.72$ / $22.07 \pm 37.70$ & $11.23 \pm 19.44$ / $10.29 \pm 17.82$ & $12.25 \pm 20.28$ / $13.24 \pm 21.34$ & $5.18 \pm 8.93$ / $4.73 \pm 8.20$ & $10.65 \pm 17.73$ / $12.13 \pm 21.01$ \\
\bottomrule
\end{tabular}%
}
\end{table}

\paragraph{Ablation and Horizon Tables.}
Table~\ref{tab:app_online_ablations} reports both the Round-12 ($384$ steps at $p=4$) and Round-16 ($512$ steps at $p=4$) endpoints for the component factorial, reuse-budget sweep, and trust-radius sweep in the 8-GPU ablation suite (with Round~16 plotted in Figure~\ref{fig:brat_online_ablations}a--c). Across Round~12 to Round~16 in Table~\ref{tab:app_online_ablations}a, unclipped \texttt{Centered + Linear} drops from $58.80\%$ to $20.13\%$ on MATH-500 as off-policy drift accumulates, while \textbf{BRAT} (\texttt{SymMax + DS}) improves across all five metrics and pulls ahead of \texttt{Centered + DS} at Round~16 on MATH-500 ($79.87\%$ vs.\ $79.73\%$), OlympiadBench ($44.81\%$ vs.\ $44.02\%$), Minerva-Math ($32.72 \pm 0.37\%$ vs.\ $29.90 \pm 2.09\%$, $+2.82$~pp), and competition Avg@16 ($25.08\%$ vs.\ $24.03\%$, $+1.05$~pp). In Table~\ref{tab:app_online_ablations}b, increasing reuse from $p=2$ to $p=4$ causes \texttt{SymMax + Linear} to collapse; AsymRE attains higher MATH-500 ($81.07\%$) and competition Avg@16 ($27.32\%$) at $p=4$, but drops from $p=2$ to $p=4$ at Round~16 on OlympiadBench ($44.31\% \to 42.43\%$) and competition Pass@16 ($50.83\% \to 48.33\%$), whereas \textbf{BRAT} improves monotonically across all five metrics from $p=1$ to $p=4$. Table~\ref{tab:app_token_gates} reports the distributed 8-GPU 16-round token-weighting comparison corresponding to Figure~\ref{fig:brat_online_ablations}d evaluated at Round~12 ($384$ updates) and Round~16 ($512$ updates) across shared seeds $\{20261021, 20261022, 20261024\}$: over the final four rounds, \textbf{BRAT} is the only compared scheme that maintained mean competition Pass@16 from Round~12 to Round~16 ($+0.55 \pm 1.40$~pp, reaching $51.85 \pm 2.36\%$), whereas \texttt{SymMax + SymDS} and both \texttt{SAPO} variants declined in mean competition Pass@16 (with Seed~1 of \texttt{SymMax + SAPO} dropping from $50.83\%$ to $40.56\%$).

\begin{figure}[!t]
\centering
\includegraphics[width=\textwidth]{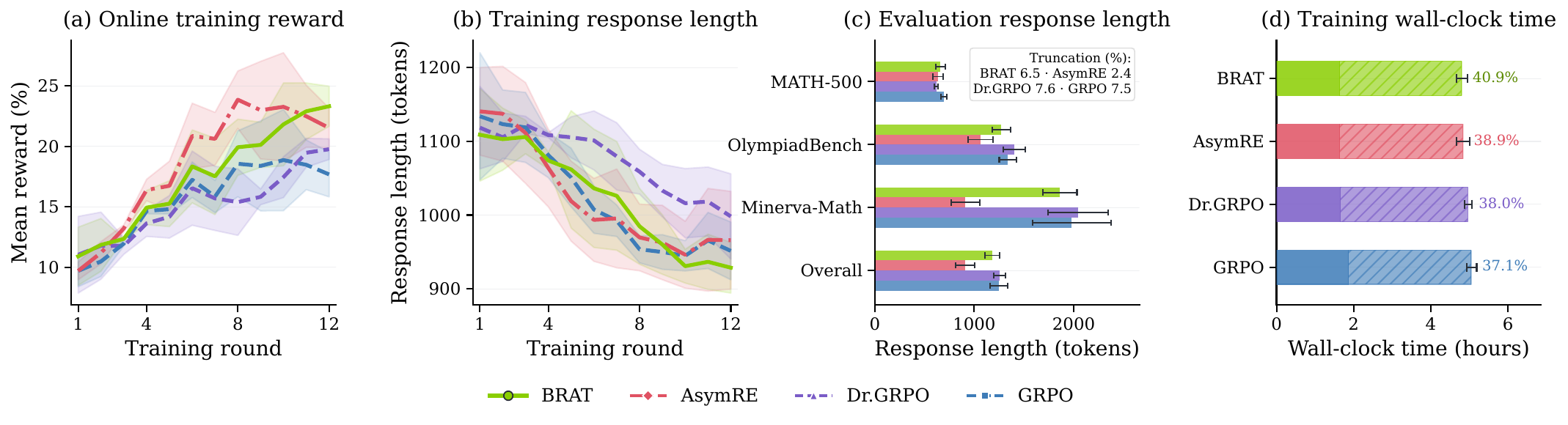}
\caption{\textbf{Training dynamics, response lengths, and wall-clock time across 12 online rounds (3 seeds).} \textbf{(a)} Online rollout reward (\%) across rounds. \textbf{(b)} Mean online rollout response length (tokens). \textbf{(c)} Greedy evaluation response length (tokens) and truncation rates across benchmarks. \textbf{(d)} Single-GPU 12-round wall-clock time decomposed into rollout generation (solid) and actor update (hatched), annotated with final Macro-Avg6.}
\label{fig:brat_dynamics}
\end{figure}

\begin{figure}[!t]
\centering
\includegraphics[width=\textwidth]{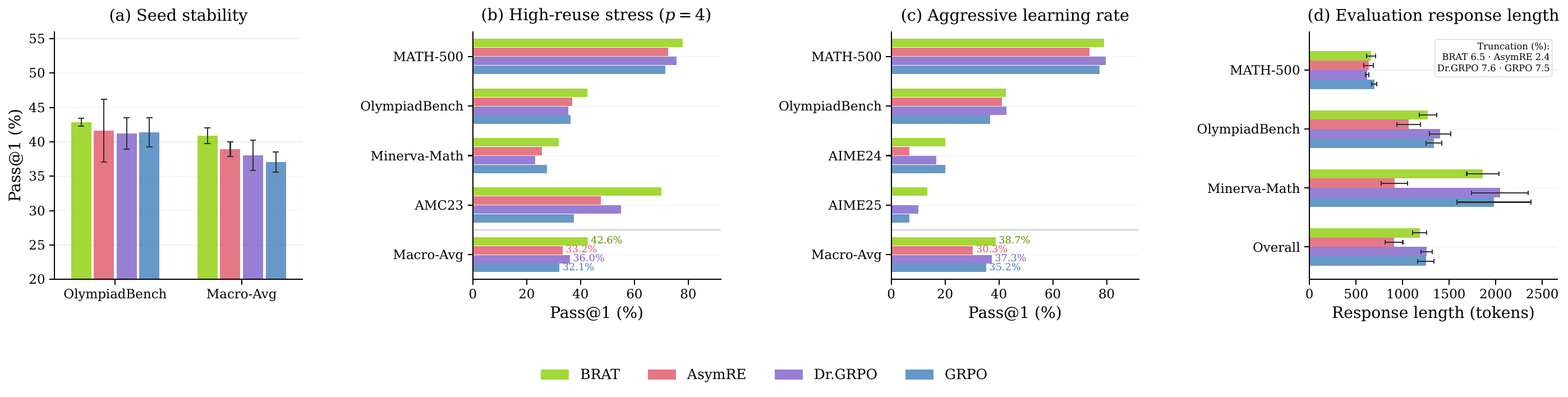}
\caption{\textbf{Extended seed stability, off-policy stress profiles, and evaluation lengths.} \textbf{(a)} Cross-seed mean and SD on OlympiadBench and Macro-Avg6. \textbf{(b)} High-reuse small-batch stress ($M=32, p=4$). \textbf{(c)} Aggressive learning-rate stress ($2.8\times 10^{-5}$). \textbf{(d)} Greedy evaluation response lengths and truncation rates.}
\label{fig:app_stress}
\end{figure}

\paragraph{Per-Seed Breakdown, Non-Truncated Accuracy, Paired Differences, and Runtime.}
Table~\ref{tab:app_per_seed} reports per-seed benchmark scores in the 12-round single-GPU main setting, and Table~\ref{tab:app_length_time} alongside Figures~\ref{fig:brat_dynamics}--\ref{fig:app_stress} summarizes online reward progression, evaluation response lengths, single-GPU wall-clock time, and off-policy stress tests ($M=32, T=16, p=4$ small-batch reuse and $2.8\times 10^{-5}$ learning rate). In Table~\ref{tab:app_length_time}, AsymRE's negative baseline shift ($b_{\mathrm{asym}}=-0.05$) is accompanied by shorter greedy responses and lower $8192$-token truncation ($2.35\%$ vs.\ $6.53\%$ for BRAT), alongside higher temperature-$1.0$ sampled competition scores at Round~12 despite lower deterministic greedy accuracy and larger variance across seeds. To verify that accuracy differences are not an artifact of truncation, evaluating only non-truncated responses across the $1{,}546$ problems yields $55.18\%$ accuracy for BRAT versus $52.03\%$ for AsymRE, and on the common subset of $1{,}300$ problems where neither method truncates on any seed, BRAT exceeds AsymRE by $+2.00$~pp. Conditioned on the three trained seed checkpoints (averaging problem-level correctness across the three seeds and forming paired problem-level $95\%$ intervals $\bar{d} \pm 1.96 s_d/\sqrt{1546}$), BRAT achieves higher mean accuracy across all $1{,}546$ problems on Micro-Avg ($+1.98$~pp over Dr.GRPO, $95\%$ CI $[+0.91, +3.06]$; $+1.66$~pp over GRPO, $95\%$ CI $[+0.54, +2.78]$; $+0.86$~pp over AsymRE, $95\%$ CI $[-0.27, +1.99]$, rising to $+1.57$~pp in the 16-round 8-GPU suite) and across the $1{,}046$ non-MATH-500 problems ($+1.66$~pp over GRPO; $+2.26$~pp over Dr.GRPO; $+0.80$~pp over AsymRE).

\begin{table}[!t]
\centering
\caption{\textbf{Distributed 8-GPU 16-round token-weighting comparison on \texttt{Qwen3-8B-Base} ($\text{mean} \pm \text{SD}$ in \% across 3 seeds at $p=4$, corresponding to Figure~\ref{fig:brat_online_ablations}d).} Bold marks the highest mean within each horizon (Round 12 and Round 16); the final column reports paired per-seed Round $12 \to 16$ changes in Competition Pass@16 ($\Delta\text{Pass@16}$ in pp).}
\label{tab:app_token_gates}
\vspace{2pt}
\setlength{\tabcolsep}{3.2pt}
\resizebox{\textwidth}{!}{%
\begin{tabular}{llccccccc}
\toprule
\textbf{Configuration (8-GPU Suite)} & \textbf{Horizon} & \textbf{Steps} & \textbf{MATH-500} & \textbf{OlympiadBench} & \textbf{Minerva-Math} & \textbf{Comp.\ Avg@16} & \textbf{Comp.\ Pass@16} & $\boldsymbol{\Delta\text{Pass@16}_{\mathrm{R12}\to\mathrm{R16}}}$ \\
\midrule
\multirow{2}{*}{\brat\textbf{ (\texttt{SymMax + DS}, $\delta=0.8$)}}
 & Round 12 & 384 & $79.27 \pm 1.03$ & $43.62 \pm 0.39$ & $29.41 \pm 0.64$ & $23.32 \pm 0.88$ & $51.30 \pm 0.98$ & \multirow{2}{*}{$\mathbf{+0.55 \pm 1.40}$} \\
 & Round 16 & 512 & $79.87 \pm 1.62$ & $44.81 \pm 1.51$ & $32.72 \pm 0.37$ & $25.08 \pm 0.39$ & $\mathbf{51.85 \pm 2.36}$ & \\
\cmidrule(lr){1-9}
\multirow{2}{*}{\texttt{SymMax + SymDS} ($\delta=0.8$)}
 & Round 12 & 384 & $78.00 \pm 1.78$ & $43.57 \pm 2.53$ & $28.06 \pm 0.56$ & $19.45 \pm 2.11$ & $51.67 \pm 1.73$ & \multirow{2}{*}{$-0.93 \pm 1.53$} \\
 & Round 16 & 512 & $78.93 \pm 0.31$ & $44.76 \pm 1.59$ & $29.41 \pm 0.97$ & $22.47 \pm 1.96$ & $50.74 \pm 3.24$ & \\
\cmidrule(lr){1-9}
\multirow{2}{*}{\texttt{Std + SAPO} ($\tau_+=1.0, \tau_-=1.05$)}
 & Round 12 & 384 & $79.60 \pm 1.11$ & $43.57 \pm 2.11$ & $31.37 \pm 5.33$ & $25.09 \pm 2.79$ & $51.20 \pm 2.94$ & \multirow{2}{*}{$-1.76 \pm 1.63$} \\
 & Round 16 & 512 & $\mathbf{80.27 \pm 0.42}$ & $43.03 \pm 0.83$ & $30.51 \pm 6.05$ & $\mathbf{26.01 \pm 1.58}$ & $49.44 \pm 1.69$ & \\
\cmidrule(lr){1-9}
\multirow{2}{*}{\texttt{SymMax + SAPO} ($\tau_+=1.0, \tau_-=1.05$)}
 & Round 12 & 384 & $\mathbf{79.73 \pm 1.42}$ & $\mathbf{46.34 \pm 1.88}$ & $\mathbf{34.07 \pm 0.42}$ & $\mathbf{26.50 \pm 1.75}$ & $\mathbf{51.76 \pm 2.10}$ & \multirow{2}{*}{$-4.91 \pm 5.72$} \\
 & Round 16 & 512 & $79.73 \pm 1.40$ & $\mathbf{46.64 \pm 1.24}$ & $\mathbf{35.42 \pm 2.09}$ & $24.55 \pm 4.15$ & $46.85 \pm 5.63$ & \\
\bottomrule
\end{tabular}%
}
\end{table}

\begin{table}[!t]
\centering
\caption{\textbf{Per-seed Greedy Pass@1 (\%) across all six mathematical reasoning benchmarks in the 12-round main setting (3 seeds).}}
\label{tab:app_per_seed}
\vspace{2pt}
\resizebox{\textwidth}{!}{%
\begin{tabular}{llcccccccc}
\toprule
\textbf{Method} & \textbf{Seed} & \textbf{MATH-500} & \textbf{OlympiadBench} & \textbf{Minerva-Math} & \textbf{AMC23} & \textbf{AIME24} & \textbf{AIME25} & \textbf{Macro-Avg6} & \textbf{Micro-Avg} \\
\midrule
\multirow{3}{*}{GRPO}
 & Seed 1 & 77.40 & 39.02 & 25.00 & 55.00 & 10.00 & 10.00 & 36.07 & 48.25 \\
 & Seed 2 & 79.60 & 41.99 & 30.88 & 50.00 & 13.33 & 16.67 & 38.75 & 51.36 \\
 & Seed 3 & 76.80 & 43.18 & 29.78 & 55.00 & 3.33 & 10.00 & 36.35 & 50.58 \\
\midrule
\multirow{3}{*}{Dr.GRPO}
 & Seed 1 & 77.80 & 43.77 & 25.37 & 55.00 & 16.67 & 20.00 & 39.77 & 50.84 \\
 & Seed 2 & 79.00 & 40.36 & 25.74 & 57.50 & 13.33 & 16.67 & 38.77 & 49.74 \\
 & Seed 3 & 77.80 & 39.47 & 26.10 & 50.00 & 10.00 & 10.00 & 35.56 & 48.64 \\
\midrule
\multirow{3}{*}{AsymRE}
 & Seed 1 & 79.00 & 45.55 & 33.09 & 60.00 & 16.67 & 6.67 & 40.16 & 53.23 \\
 & Seed 2 & 79.00 & 42.73 & 29.41 & 52.50 & 16.67 & 10.00 & 38.38 & 51.23 \\
 & Seed 3 & 77.80 & 36.65 & 27.57 & 67.50 & 10.00 & 10.00 & 38.25 & 48.12 \\
\midrule
\multirow{3}{*}{\brat\textbf{ (Ours)}}
 & Seed 1 & 80.00 & 43.47 & 30.15 & 62.50 & 16.67 & 20.00 & 42.13 & 52.46 \\
 & Seed 2 & 79.20 & 42.43 & 28.31 & 60.00 & 16.67 & 16.67 & 40.55 & 51.29 \\
 & Seed 3 & 79.60 & 42.58 & 28.31 & 62.50 & 13.33 & 13.33 & 39.94 & 51.42 \\
\bottomrule
\end{tabular}%
}
\end{table}

\begin{table}[!t]
\centering
\caption{\textbf{Generation length, truncation rate, and single-GPU wall-clock training time across 12 online rounds ($\text{mean} \pm \text{SD}$ over 3 seeds).} Evaluation lengths and truncation rates (at $8192$ generated tokens) are aggregated across all $1{,}546$ problems per seed.}
\label{tab:app_length_time}
\vspace{2pt}
\resizebox{\textwidth}{!}{%
\begin{tabular}{lcccccccc}
\toprule
\textbf{Method} & \textbf{Final Online} & \textbf{Final Online} & \textbf{Eval Length} & \textbf{Eval Trunc.\ Rate} & \textbf{Rollout Time} & \textbf{Actor Time} & \textbf{Total Time} & \textbf{Min /} \\
 & \textbf{Reward (\%)} & \textbf{Length (tok)} & \textbf{All6 (tok)} & \textbf{All6 (\%)} & \textbf{(hours)} & \textbf{(hours)} & \textbf{(hours)} & \textbf{Round} \\
\midrule
GRPO & $19.47 \pm 3.23$ & $956.8 \pm 90.5$ & $1248.1 \pm 88.3$ & $7.46 \pm 0.97$ & $1.86 \pm 0.05$ & $3.18 \pm 0.09$ & $5.06 \pm 0.13$ & $25.3$ \\
Dr.GRPO & $20.05 \pm 1.80$ & $1014.7 \pm 40.2$ & $1256.0 \pm 60.0$ & $7.63 \pm 0.70$ & $1.65 \pm 0.05$ & $3.30 \pm 0.05$ & $4.97 \pm 0.10$ & $24.8$ \\
AsymRE & $22.59 \pm 3.14$ & $979.4 \pm 111.6$ & $906.0 \pm 95.4$ & $2.35 \pm 0.84$ & $1.63 \pm 0.07$ & $3.20 \pm 0.12$ & $4.84 \pm 0.18$ & $24.2$ \\
\brat\textbf{ (Ours)} & $\mathbf{25.20 \pm 2.40}$ & $\mathbf{948.7 \pm 60.2}$ & $1183.3 \pm 74.6$ & $6.53 \pm 0.76$ & $\mathbf{1.63 \pm 0.06}$ & $\mathbf{3.17 \pm 0.09}$ & $\mathbf{4.81 \pm 0.15}$ & $\mathbf{24.0}$ \\
\bottomrule
\end{tabular}%
}
\end{table}

\end{document}